\documentclass{article}

\usepackage[utf8]{inputenc} % allow utf-8 input
\usepackage[T1]{fontenc}    % use 8-bit T1 fonts
\usepackage{url}            % simple URL typesetting
\usepackage{booktabs}       % professional-quality tables
\usepackage{amsfonts}       % blackboard math symbols
\usepackage{nicefrac}       % compact symbols for 1/2, etc.
\usepackage{microtype}      % microtypography
\usepackage{xcolor}         % colors

\usepackage{graphicx} % Required for inserting images

\usepackage[colorlinks,linkcolor=red, anchorcolor=blue, citecolor=blue]{hyperref}

\usepackage{amsmath, amssymb, amsthm}
\usepackage{geometry}
\usepackage{booktabs}
\usepackage{xcolor}
\usepackage{hyperref}
\usepackage{bm}
\usepackage{natbib}
\usepackage{relsize}
\usepackage{xspace}

\usepackage{listings}  % Code

\newcommand{\methodShort}{Cubit\xspace}
\newcommand{\method}{Limited-Range Rescale\xspace}
\newcommand{\methodS}{LRR\xspace}

\title{Cubit: Token Mixer with Kernel Ridge Regression}

\author{%
  Chuanyang Zheng, Jiankai Sun, Yihang Gao, Yuehao Wang, Liangchen Tan, \\ Mac Schwager, Anderson Schneider, Yuriy Nevmyvaka, Xiaodong Liu \\
  Contact: cyzhengme@gmail.com
}

\begin{document}

\maketitle

\begin{abstract}
Since its introduction in 2017, the Transformer has become one of the most widely adopted architectures in modern deep learning. 
Despite extensive efforts to improve positional encoding, attention mechanisms, and feed-forward networks, the core token-mixing mechanism in Transformers remains attention.
In this work, we show that the attention module in Transformers can be interpreted as performing Nadaraya–Watson regression, where it computes similarities between tokens and aggregates the corresponding values accordingly.
Motivated by this perspective, we propose \textbf{Cubit,  a potential next-generation architecture that leverages Kernel Ridge Regression (KRR), while the vanilla Transformer relies on Nadaraya-Watson regression.} 
Specifically, Cubit modifies the classical attention computation by incorporating the closed-form solution of KRR, combining value aggregation through kernel similarities with normalization via the inverse of the kernel matrix.
To improve the training stability, we further propose the \textbf{Limited-Range Rescale (LRR)}, which rescales the value layer within a controlled range.
We argue that Cubit, as a KRR-based architecture, provides a stronger mathematical foundation than the vanilla Transformer, whose attention mechanism corresponds to Nadaraya–Watson regression. We validate this claim through comprehensive experiments. 
The experimental results suggest that Cubit may exhibit stronger long-sequence modeling capability. In particular, its performance gain over the Transformer appears to increase as the training sequence length grows.
\end{abstract}

\section{Introduction}

Recurrent Neural Networks (RNNs), introduced in the 1980s \cite{hopfield1982neural,jordan1986serial,elman1991distributed,graves2012long}, process sequences by recurrently updating hidden states across tokens, incurring linear computational complexity with respect to sequence length. In 2017, the Transformer architecture \cite{vaswani2017attention} was introduced, proposing self-attention mechanisms that capture global dependencies but at the quadratic computational cost. Despite this scalability limitation, Transformers have become the dominant architecture across natural language processing, computer vision, and multimodal learning, achieving remarkable empirical success. Using the powerful Transformer architecture, significant progress has been made in language modeling \cite{fedus2022switch,puigcerver2023sparse,jiang2024mixtral,meta2025llama,liu2024deepseek,team2025kimi} and computer vision \cite{riquelme2021scaling,lin2023video}. 
The mixture-of-experts architecture \cite{jacobs1991adaptive,shazeer2017outrageously,roller2021hash} has emerged as an efficient alternative that allows parameter scaling while maintaining manageable computational requirements. 
More recently, combining Transformer backbones with mixture-of-experts designs has enabled the development of extremely large yet computationally efficient language models, demonstrating the effectiveness of sparse parameter scaling \cite{dai2024deepseekmoe,jiang2024mixtral,shen2024jetmoe,wei2024skywork,liu2024deepseek}.

The Transformer architecture can be theoretically grounded in Nadaraya-Watson Regression \cite{nadaraya1964estimating,watson1964smooth} . The architecture comprises two principal components: feed-forward networks (FFNs) and multi-head self-attention mechanisms. The FFN layers can be interpreted as key-value memory systems \cite{geva2021transformer}. The attention mechanism itself operates as Nadaraya-Watson Regression with an exponential kernel (via softmax) and L1 normalization. Consequently, despite numerous architectural modifications were proposed—including sparse attention, linear attention, and various approximation schemes—the fundamental computational paradigm remains rooted in Nadaraya-Watson regression with inherent quadratic complexity.

In this work, we introduce Cubit, which replaces the Nadaraya-Watson regression in the attention module of Transformers with Kernel Ridge Regression \cite{murphy2012machine,williams1995gaussian}. 
% The Cubit layer computes:
%   $$\text{Softmax}(Q^TK)(\text{Softmax}(K^T\text{Norm}(K)))^{-1}V,$$
%   while the $Q$, $K$ and $V$ are the query embedding, key embedding, and value embedding. 
% KRR exhibits strong connections to Kernel Ridge Regression \cite{murphy2012machine}, sharing analogous predictive formulations. 
This framework readily extends to alternative regression methodologies, including local linear regression variants \cite{macaulay1931introduction,cleveland2013smoothing,murray2019wf}.
Compared to Transformers based on Nadaraya-Watson Regression, Cubit with Kernel Ridge Regression offers several theoretical benefits, including explicit regularization for bias-variance trade-off, faster convergence rates in RKHS, and greater robustness to noise and boundary effects \cite{long2024optimal,bak2025effect,mollenhauer2025regularized,wen2025robustness,barzilai2025overfitting}.
We summarize our principal contributions as follows:
\begin{itemize}
\item We establish a unified theoretical framework connecting token mixing mechanisms to classical regression methods, systematically analyzing \textbf{Nadaraya-Watson Regression and Kernel Ridge Regression}. This perspective helps to understand and develop novel model architectures.
\item Compared to  Transformer based on Nadaray-Watson Regression), we propose the \textbf{Cubit based on Kernel Ridge Regression}, with \textbf{\method(\methodS)} to improve the training stability.
\item  We validate \methodShort across diverse datasets, sequence lengths, and model sizes. The experimental results suggest that \methodShort exhibits stronger long-sequence modeling capability than the Transformer. In particular, its performance gain tends to increase as the training sequence length becomes longer.

\end{itemize}

\section{Related Work}

% skip connection, dense connection, hyper connection, mhc, virtual width network, geonorm, and so on
% \paragraph{Skip Connection}

% Linear; Quadratic
% \paragraph{Attention Manicheism}

% process sequence

\paragraph{Transformer Architecture} Transformer architecture \cite{vaswani2017attention} was proposed in 2017, with Feed-Forward Network and Attention. In the following years, there are modifications of Transformer. From the view of FFN, there are mixture-of-expert and different activation functions. From the view of attention, there are GQA \cite{ainslie2023gqa}, MQA \cite{shazeer2019fast}, MLA \cite{liu2024deepseek}, TPA \cite{zhang2025tensor} and so on. There is also gated attention \cite{qiu2025gated} to improve the performance. Also, there are works that are trying to replace the softmax attention with ReLU attention \cite{wortsman2023replacing} and sigmoid attention \cite{ramapuram2024theory}. The skip connection \cite{he2016deep} is also discussed, such as hyper-connection \cite{zhu2024hyper}, attention residual \cite{team2026attention}, deepnorm \cite{wang2024deepnet}, and sandwitchnorm \cite{ding2021cogview}. However, these modifications do not modify the modeling of the attention block, which is actually the Nadaraya-Watson Regression.

\paragraph{Token Mixer}  To address the scalability limitations of quadratic attention, numerous linear-complexity linear token mixers have been developed. Early recurrent approaches include LSTM \cite{graves2012long} and its extension mLSTM \cite{beck2024xlstm} with matrix-valued memory states. Linear attention mechanisms \cite{kasai2021finetuning} approximate softmax attention via kernel feature mappings to achieve linear complexity.
State space models (SSMs) represent another major family, beginning with S4 \cite{gu2021efficiently} and its variants \cite{gu2022parameterization}. Mamba \cite{gu2024mamba} introduced input-dependent selective state updates with hardware-aware parallel scans, later unified with attention via SSD in Mamba-2 \cite{dao2024transformers}. Linear RNN variants include RetNet \cite{sun2023retentive}, RWKV \cite{peng2025rwkv}, HGRN-2 \cite{qin2024hgrn2}, and Gated DeltaNet \cite{yang2024gated}. MegaLodon \cite{ma2024megalodon} combines exponential moving average with chunked attention. MLP-Mixer \cite{tolstikhin2021mlp} demonstrates token mixing without recurrence or attention via channel-wise MLPs. Transformer \cite{vaswani2017attention,xu2025deltaformer} employs self-attention with quadratic complexity, enabling strong relational reasoning and in-context learning at the cost of computational scalability for long sequences.

\paragraph{Regression Method} Regression methods fall into three categories by learning approach. Non-parametric methods make minimal functional assumptions, letting data determine model structure. Key examples include k -nearest neighbor regression \cite{fix1985discriminatory}, kernel-based approaches (Nadaraya-Watson \cite{nadaraya1964estimating}, ridge kernel \cite{murphy2012machine}, and local linear/polynomial variants \cite{cleveland1981lowess, cleveland1988locally}), and Kernel Ridge Regression \cite{williams1995gaussian}—which defines a function-space prior via covariance kernels for predictive distributions with built-in uncertainty.
Parametric methods assume fixed functional forms with finite parameters: linear regression \cite{freedman2009statistical,berk2004regression} for interpretable relationships, polynomial regression \cite{theil1950rank} for non-linear patterns, and logistic regression \cite{tolles2016logistic} for categorical responses via log-odds modeling.
Semi-parametric methods balance both worlds: partial linear models \cite{engle1986semiparametric,zeger1994semiparametric} combine linear parametric terms with non-parametric components, while Generalized Additive Models \cite{nelder1972generalized} represent responses as sums of smooth univariate functions, preserving additive structure while capturing complex patterns.

% The predictive variance:
% \begin{equation}
%     \bm{\sigma}^{2,(h)} = \text{diag}\left(\bm{K}^{(h)} - \bm{K}^{(h)} \bm{\Sigma}^{(h)} \bm{K}^{(h)\top}\right) \in \mathbb{R}^{N},
% \end{equation}
% quantifies aggregation uncertainty for each token. High variance indicates regions where the kernel similarity provides weak evidence—typically occurring at sequence boundaries or semantic discontinuities. This uncertainty signal enables targeted interventions, such as adaptive bandwidth adjustment or auxiliary boundary processing, directly addressing the boundary bias inherent in fixed-kernel regression.

\section{Method}

\subsection{Kernel Ridge Regression}

We begin by revisiting the kernel ridge regression (KRR) method, which provides key insights into interpreting attention mechanisms as regression procedures. This perspective further motivates the design of a novel attention mechanism grounded in the KRR formulation.

Consider a positive definite kernel $K: \mathcal{X} \times \mathcal{X} \rightarrow \mathbb{R}$ and its associated reproducing kernel Hilbert space $\mathcal{H}_K$ with norm $\|\cdot\|_{\mathcal{H}_K}$. By the Moore-Aronszajn theorem, such a kernel uniquely defines an RKHS where evaluation functionals are continuous. The representation theorem states that the minimizer of the regularized empirical risk over $\mathcal{H}_K$ admits a finite-dimensional representation in terms of kernel evaluations at training points. More specifically, we consider the problem of learning a vector-valued function $f: \mathcal{X} \to \mathbb{R}^m$ from a given dataset $\{(\bm{x}_i, \bm{y}_i)\}_{i=1}^N$, where $\bm{x}_i \in \mathcal{X} \subseteq \mathbb{R}^{d}$ and $\bm{y}_i \in \mathbb{R}^m$. KRR solves the following optimization problem:
\begin{equation}
\label{eq1}
    \min_{f \in \mathcal{H}_K} \sum_{i=1}^N \left\|\bm{y}_i - f(\bm{x}_i)\right\|_{2}^2 + \lambda \|f\|_{\mathcal{H}_K}^2,
\end{equation}
where $\lambda > 0$ is the regularization parameter controlling the bias-variance trade-off. The representation theorem guarantees that the optimal solution takes the form:
\begin{equation}
\label{eq2}
    f(\bm{x}) = \sum_{i=1}^N \bm{c}_i K(\bm{x}, \bm{x}_i),
\end{equation}
which reduces the infinite-dimensional optimization in the functional space to finding coefficients $\bm{C} = [\bm{c}_1, \ldots, \bm{c}_N]^{\top} \in \mathbb{R}^{N \times m}$ within a finite-dimensional vector space.

Substituting the kernel expansion \eqref{eq2} into the objective \eqref{eq1} yields the finite-dimensional optimization:
\begin{equation}
\label{eq3}
    \min_{\bm{C}} \|\bm{Y} - \bm{K}\bm{C}\|_{\mathrm{F}}^2 + \lambda \, \mathrm{tr}(\bm{C}^\top \bm{K} \bm{C}),
\end{equation}
where $\bm{Y}=[\bm{y}_1,\cdots,\bm{y}_{N}]^{\top} \in \mathbb{R}^{N \times m}$ denotes the collection of label vectors over all samples, and $\bm{K} \in \mathbb{R}^{N \times N}$ is the Gram (kernel) matrix with entries $K_{ij} = K(\bm{x}_i, \bm{x}_j)$. 
It is worthnoting that \eqref{eq3} is convex problem in $\bm{C}$, the global solution admits the following linear system due to the stationary condition:
\begin{equation}
   (\bm{K} + \lambda \bm{I}) \bm{C} = \bm{Y},
\end{equation}
where $\bm{I} \in \mathbb{R}^{N \times N}$ is the identity matrix. Therefore, the optimal coefficient matrix is given by
\begin{equation*}
    \bm{C} = (\bm{K} + \lambda \bm{I})^{-1} \bm{Y}.
\end{equation*}
For any new input $\bm{x} \in \mathcal{X}$, define
\[
\bm{k}(\bm{x}) =
\begin{bmatrix}
K(\bm{x}, \bm{x}_1) \\
\vdots \\
K(\bm{x}, \bm{x}_N)
\end{bmatrix}
\in \mathbb{R}^N.
\]
The prediction is then given by
\begin{equation}
\label{eq4}
    f(\bm{x}) = \bm{k}(\bm{x})^\top \bm{C}
= \bm{k}(\bm{x})^\top (\bm{K} + \lambda \bm{I})^{-1} \bm{Y}.
\end{equation}
Here, $\bm{k}(\bm{x})^\top \bm{Y}$ in \eqref{eq4} can be interpreted as Nadaraya-Watson regression estimator, while $(\bm{K} + \lambda \bm{I})^{-1}$ acts as a normalization operator that couples the information across all data samples.

\subsection{Interpreting the Attention Mechanism as Regression}

We interpret the token mixing operation in the attention block of the Transformer as a form of similarity-based regression (Nadaraya-Watson Regression \cite{nadaraya1964estimating}), where each token aggregates information from other tokens according to learned similarity scores.

Given $N$ token embeddings $\bm{X} \in \mathbb{R}^{N \times D}$, we compute the query, key, and value representations for each head $h \in \{1,2,\cdots, H\}$:
\begin{equation*}
    \bm{Q}^{(h)} = \bm{X} \bm{W}_Q^{(h)}, \quad 
    \bm{K}^{(h)} = \bm{X} \bm{W}_K^{(h)}, \quad 
    \bm{V}^{(h)} = \bm{X} \bm{W}_V^{(h)},
\end{equation*}
where $\bm{W}_Q^{(h)}, \bm{W}_K^{(h)}, \bm{W}_V^{(h)} \in \mathbb{R}^{D \times d_h}$ are learnable projection matrices with $d_h = D/H$.
The attention weights are computed via scaled dot-product similarity with a learnable temperature parameter $w^{(h)} > 0$:
\begin{equation}
\label{eq6}
    \bm{A}^{(h)} = \text{Softmax}\left( w^{(h)} \cdot\bm{Q}^{(h)} \bm{K}^{(h)\top}\right),
\end{equation}
where $\text{Softmax}(\cdot)$ is applied row-wise to ensure normalization.
Then, the output of each head is given by
\begin{equation}
\label{eq5}
    \bm{Z}^{(h)} = \bm{A}^{(h)} \bm{V}^{(h)} 
    = \text{Softmax}\left(w^{(h)} \cdot \bm{Q}^{(h)} \bm{K}^{(h)\top}\right) \bm{V}^{(h)}.
\end{equation}

From \eqref{eq5}, we observe that each output token is computed as a weighted aggregation of value vectors, where the weights are determined by query-key similarity scores in \eqref{eq6}. In particular, the $i$-th output token can be written as
\[
\bm{z}_i^{(h)} = \sum_{j=1}^N A_{ij}^{(h)} \bm{v}_j^{(h)},
\]
where $A_{ij}^{(h)}$ denotes the attention weight between token $i$ and token $j$.
% \paragraph{Connection to kernel regression.}
% This aggregation structure closely resembles kernel-based regression methods. In particular, in Nadaraya--Watson kernel regression, the prediction at a query point $\bm{x}$ is given by
% \[
% f(\bm{x}) = \sum_{i=1}^N \tilde{K}(\bm{x}, \bm{x}_i)\,\bm{y}_i,
% \]
% where $\tilde{K}(\bm{x}, \bm{x}_i)$ is the normalized kernel weight function.
% Comparing this with \eqref{eq5}, we observe that
% \begin{itemize}
%     \item The attention weights $A_{ij}^{(h)}$ play the role of normalized kernel similarities;
%     \item The query-key dot product $\text{Softmax}\left(\bm{Q}^{(h)\top} \bm{K}^{(h)}\right)$ acts as a learned similarity function;
%     \item The value vectors $\bm{V}^{(h)}$ correspond to the regression targets being aggregated.
% \end{itemize}
Therefore, the attention mechanism can be interpreted as a form of data-dependent kernel regression, where both the similarity function and the regression targets are learned from data.

\paragraph{Difference from kernel ridge regression.}
While the structural form is similar in output aggregation, attention differs from KRR in the construction of the weights, comparing \eqref{eq4} with \eqref{eq5}. In KRR, the prediction involves solving a global linear system with the inverse Gram matrix $(\bm{K} + \lambda \bm{I})^{-1}$, which introduces an explicit regularization effect. In contrast, attention employs the normalization via the softmax operator, resulting in adaptive, query-dependent weights without solving a global system.

\paragraph{Adaptive bandwidth.}
The learnable parameter $w^{(h)}$ controls the sharpness of the distribution. As $w^{(h)} \to 0$, the weights approach uniform averaging, corresponding to global aggregation. As $w^{(h)} \to \infty$, the weights concentrate on the most similar tokens, resembling nearest-neighbor regression. This allows different heads to capture interactions at multiple scales.

\paragraph{Summary.}
In summary, the attention mechanism can be viewed as a form of similarity-based regression, closely related to kernel smoothing methods such as Nadaraya--Watson regression, but with learned similarity functions.

\subsection{Designing the Attention Mechanism via Kernel Ridge Regression}

Motivated by the kernel regression interpretation of the standard attention mechanism, we propose a new attention formulation based on KRR, where both similarity evaluation and inverse normalization are explicitly incorporated into the attention scores.

\paragraph{Kernel Ridge Regression Formulation.} 
For head $h$, let $\phi(\cdot)$ denote the feature map induced by a kernel function $K(\cdot,\cdot)$. By Mercer's theorem, the kernel matrix must be constructed consistently from the same RKHS in order to ensure positive definiteness and the validity of the Representer theorem. Accordingly, the kernel function can be written as
\begin{equation*}
    K(\bm{x}_i,\bm{x}_j) = \psi(\bm{x}_i)   \psi(\bm{x}_j)^{\top},
\end{equation*}
where $\psi:\mathbb{R}^{D}\to\mathbb{R}^{d_h}$ defines the feature transformation associated with the kernel.

To adapt this framework to attention, we parameterize $\psi(\bm{x}) = \phi(\bm{x}\bm{W}_{K}^{(h)})$, where $\phi$ is a prescribed activation function and $\bm{W}_K^{(h)}$ is a trainable projection matrix. The corresponding normalization term in our kernel regression attention is defined as
\begin{equation*}
    \bm{\Sigma}^{(h)} = \left(\mathbf{K}^{(h)} \mathbf{K}^{(h)\top} + \lambda \mathbf{I}\right)^{-1},
\end{equation*}
where $\mathbf{K}^{(h)} = \phi(\bm{X} \bm{W}_K^{(h)})$. 
In practice, the kernel matrix admits various design choices. The key transformation $\phi$ can range from identity mapping to normalization operators (e.g., $\ell_2$ normalization), while the kernel activation $g(\cdot)$ can be flexibly configured. This leads to a general formulation:
\begin{equation}
    \bm{\Sigma}^{(h)} = \left(g\left(\mathbf{K}^{(h)} \mathbf{K}^{(h)\top}\right) + \lambda \mathbf{I}\right)^{-1},
\end{equation}
where different combinations of $\phi$ and $g$ yield various kernel instantiations.
where different combinations of $\phi$ and $g$ yield various kernel instantiations, while the $g$ could be $\text{Softmax}$ for both $\bm{A}^{(h)}$ and $\bm{\Sigma}^{(h)}$.
It remains to model 
\[
\bm{k}(\bm{x}_{i}) =
\begin{bmatrix}
K(\bm{x}_{i}, \bm{x}_{1}) \\
\vdots \\
K(\bm{x}_{i}, \bm{x}_{i})
\end{bmatrix}
\in \mathbb{R}^{i}.
\]
This autoregressive construction reflects the causal constraint that the $i$-th token can only attend to itself and previous tokens, so only the first $i$ samples are available in the regression.
Under the standard kernel formulation, we would have
\begin{equation*}
    \bm{k}(\bm{x}_{i}) = 
    \begin{bmatrix}
\phi(\bm{x}_{i}\bm{W}_{K}^{(h)}) \phi(\bm{x}_{1}\bm{W}_{K}^{(h)})^{\top} \\
\vdots \\
\phi(\bm{x}_{i}\bm{W}_{K}^{(h)}) \phi(\bm{x}_{i}\bm{W}_{K}^{(h)})^{\top}
\end{bmatrix}.
\end{equation*}
Although this formulation closely follows classical kernel theory, it is overly restrictive for attention modeling in practice. To improve flexibility, we adopt the query-key parameterization from standard attention and instead define
\begin{equation*}
    \bm{k}(\bm{x}_{i}) = 
    \begin{bmatrix}
\phi(\bm{x}_{i}\bm{W}_{Q}^{(h)}) \phi(\bm{x}_{1}\bm{W}_{K}^{(h)})^{\top} \\
\vdots \\
\phi(\bm{x}_{i}\bm{W}_{Q}^{(h)}) \phi(\bm{x}_{i}\bm{W}_{K}^{(h)})^{\top}
\end{bmatrix}.
\end{equation*}
This relaxation introduces greater expressive power and improves numerical stability. In particular, when $\phi$ is the identity map and softmax normalization is applied to
$[\bm{k}(\bm{x}_{i});\cdots;\bm{k}(\bm{x}_{N})]$, the resulting formulation reduces to the standard attention mechanism in \eqref{eq6}.

Therefore, the final output in \eqref{eq5} is given by
\begin{equation}
    \mathbf{Z}^{(h)} =  \mathbf{A}^{(h)}\bm{\Sigma}^{(h)}\mathbf{V}^{(h)}, 
    \label{eq:krr_output}
\end{equation}
where $\mathbf{A}^{(h)}$ denotes the kernel similarity matrix between each token and its accessible context tokens, and $\bm{\Sigma}^{(h)}$ is the KRR-inspired normalization matrix.

\paragraph{\method.} 
We introduce a learnable scaling vector $\mathbf{s}^{(h)} \in \mathbb{R}^N$ to adaptively adjust the kernel precision, where each element is computed as:
\begin{equation}
    \hat{s}_i^{(h)} = \alpha \cdot \sigma\left([\mathbf{W}_s \mathbf{x}]_i\right) + \beta, \quad \forall i \in \{1, \dots, N\},
\end{equation}
with $\sigma(\cdot)$ denoting the sigmoid function, and $\alpha, \beta > 0$ ensuring $\hat{s}_i^{(h)} \in (\beta, \alpha+\beta)$. This guarantees the invertibility of $\mathbf{\hat{S}}^{(h)} = \mathrm{diag}(\mathbf{\hat{s}}^{(h)})$, with bounded inverse elements $\frac{1}{\hat{s}_i^{(h)}} \in \left(\frac{1}{\alpha+\beta}, \frac{1}{\beta}\right)$.
The scaling mechanism modifies the kernel ridge regression solution. Specifically, we solve:
\begin{equation}
    \mathbf{\hat{S}}^{(h)} (\bm{\Sigma}^{(h)})^{-1} \mathbf{o} = \mathbf{V},
    \label{eq:scaled_krr}
\end{equation}
% which yields:
\begin{equation}
     \mathbf{o} = (\bm{\Sigma}^{(h)})(\mathbf{\hat{S}}^{(h)})^{-1}V
\end{equation}
We could have $S^{(h)}=(\hat{S}^{h})^{-1}$.
Finally, the KRR-based token mixer produces the output via the following:
\begin{equation}
    \mathbf{Z}^{(h)} = \mathbf{A}^{(h)}O= \mathbf{A}^{(h)} \bm{\Sigma}^{(h)} \mathbf{S}^{(h)} \mathbf{V}^{(h)}.
    \label{eq:krr_output}
\end{equation}

\paragraph{Local Linear Regression and Kernel Ridge Regression.} We further discuss the Cubit with Local Linear Regression Appendix \ref{appendix: llr}.

\paragraph{Differences between DeltaFormer and Cubit.} The DeltaFormer also discusses how to use the inverse matrix to improve the performance, so that we can discuss the difference here. First, the motivation is different. The DeltaFormer is motivated by combining DeltaNet and Transformer, while the Cubit is motivated by the Kernel Ridge Regression. Secondly, the implementation is different, as the DeltaFormer uses two different embeddings $w$ and $k$ to construct the inverse matrix, while Cubit suggests that we have to use one specific embedding (e.g., only $k$ or only $r$) so that the matrix is invertible. Thirdly, motivated by DeltaNet, the DeltaFormer sets the diagonal value to 1 to calculate the inverse matrix,  while the Cubit can be sure that there exists an inverse matrix so that the diagonal value does not necessarily have to be 1. Such an error may gradually increase with the sequence length increase (empirically validated in Section \ref{label:training_length}). Finally, the \methodShort proposes \method (\methodS) to improve the performance, while DeltaFormer does not have. The \methodShort implementation is in Appendix \ref{appendix: implementation}.

\section{Experiment}

\paragraph{Baseline.}
We compare the proposed \methodShort with the Transformer \cite{vaswani2017attention} and DeltaFormer \cite{xu2025deltaformer}.
The Transformer is the foundation of the recent Large Language Model. The DeltaFormer combines the DeltaNet and Transformer, suggesting that it may have higher expressiveness than the Transformer.

% \paragraph{Datasets}
% \vspace{-5pt}
\paragraph{Datasets.}
Our analysis involves training language models on the Arxiv and Books3 datasets, which are frequently used for evaluating model performance \citep{press2021train}. Moreover, we train the model on large-scale dataset FinWeb-Edu \citep{penedo2024fineweb,lozhkov2024fineweb-edu} and evaluate on downstream datasets, 
including ARC~\citep{clark2018think},  HellaSwag~\citep{zellers2019hellaswag},  PIQA \citep{bisk2020piqa}, SciQ \citep{welbl2017crowdsourcing}, and   WinoGrade~\citep{sakaguchi2021winogrande}, SocialIQA \cite{sap2019social}, and RACE \cite{lai2017race}.
% \vspace{-5pt}
\paragraph{Experiment settings.}
Initially, we compare \methodShort with other baselines at training lengths 512 and 1024, using decoder-only Transformers \citep{brown2020language} with model size 125M, whose configuration is shown in Appendix \ref{model configuration details}. Subsequently, we evaluate the performance of larger model sizes, specifically 350M and 1.3 B. 
Finally, we suggest that the \methodShort may presents advantages for long sequences.

\subsection{Compare with Baseline}

\begin{figure}[htbp]
%\begin{figure}
\centering
\includegraphics[width=0.45\textwidth]{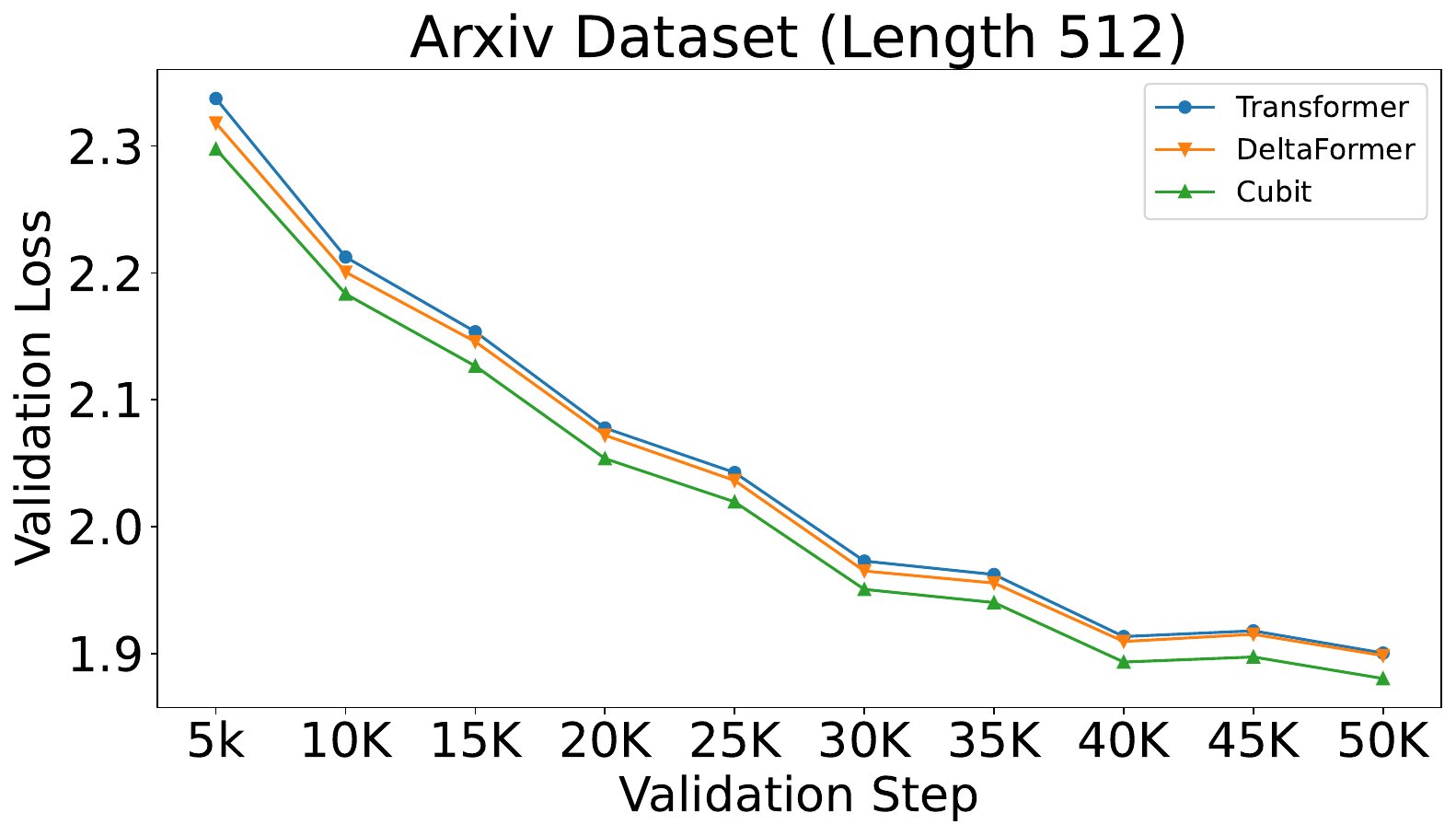}
\hspace{0in}
\includegraphics[width=0.45\textwidth]{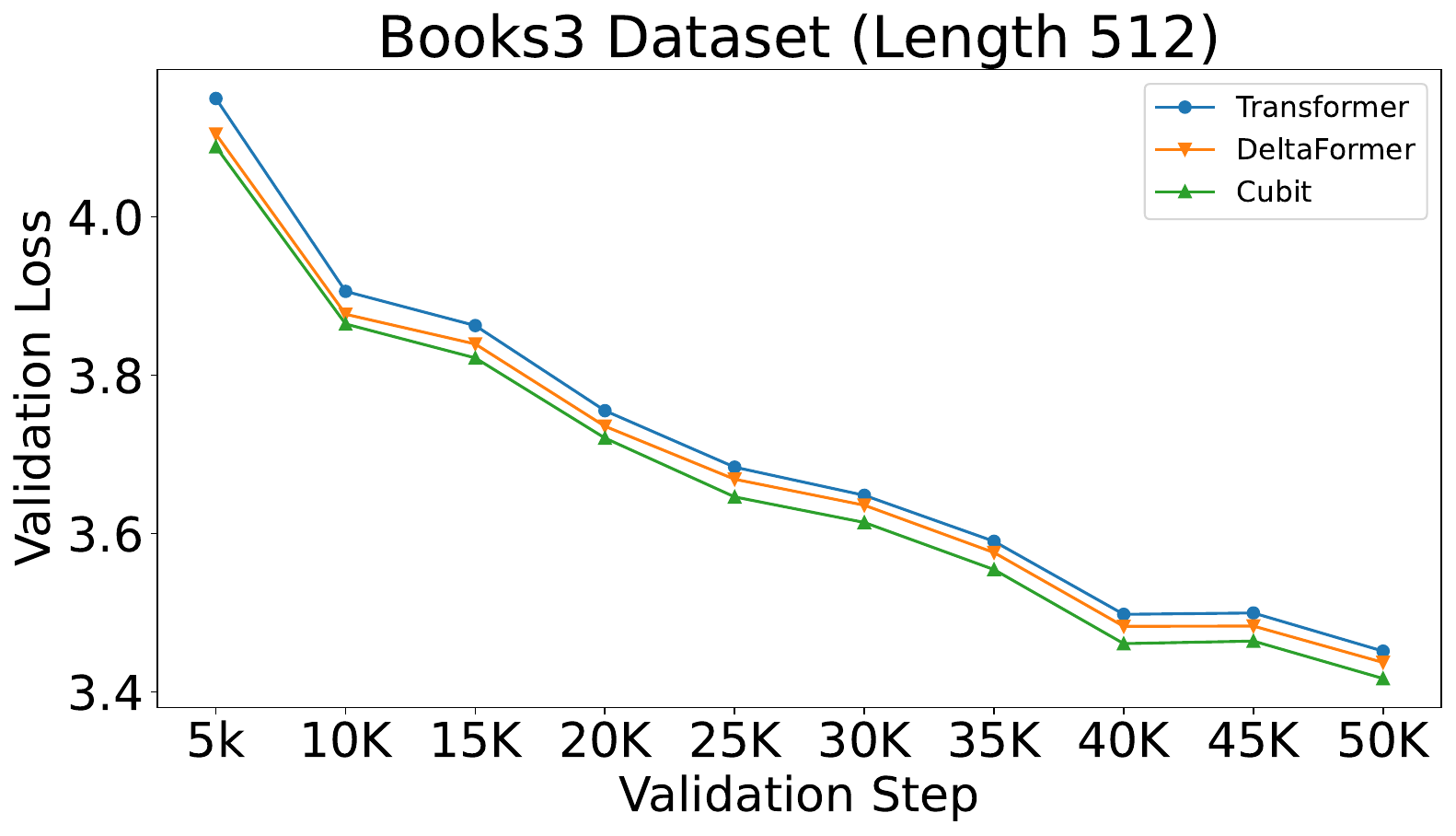}
\includegraphics[width=0.45\textwidth]{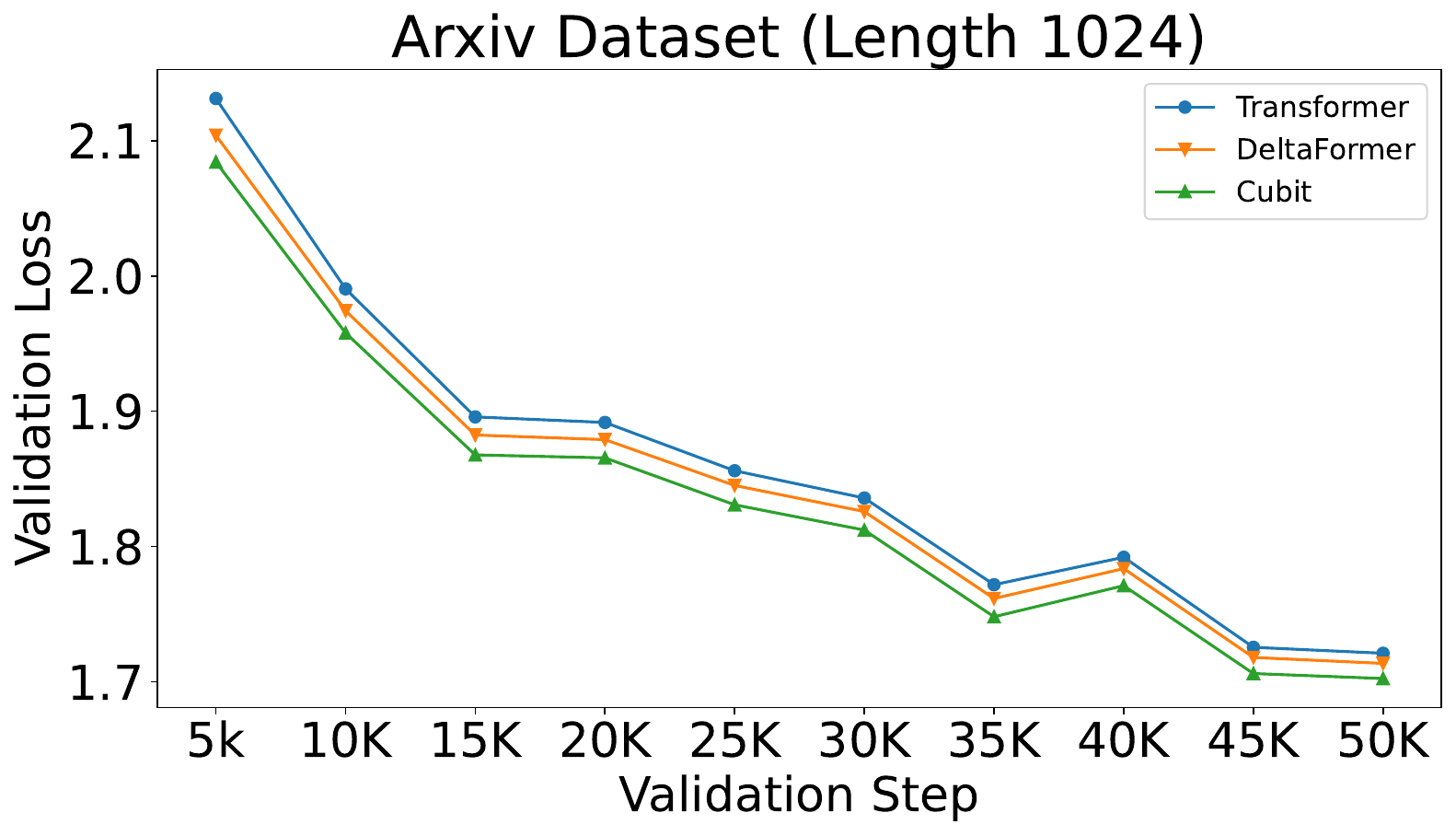}
\hspace{0in}
\includegraphics[width=0.45\textwidth]{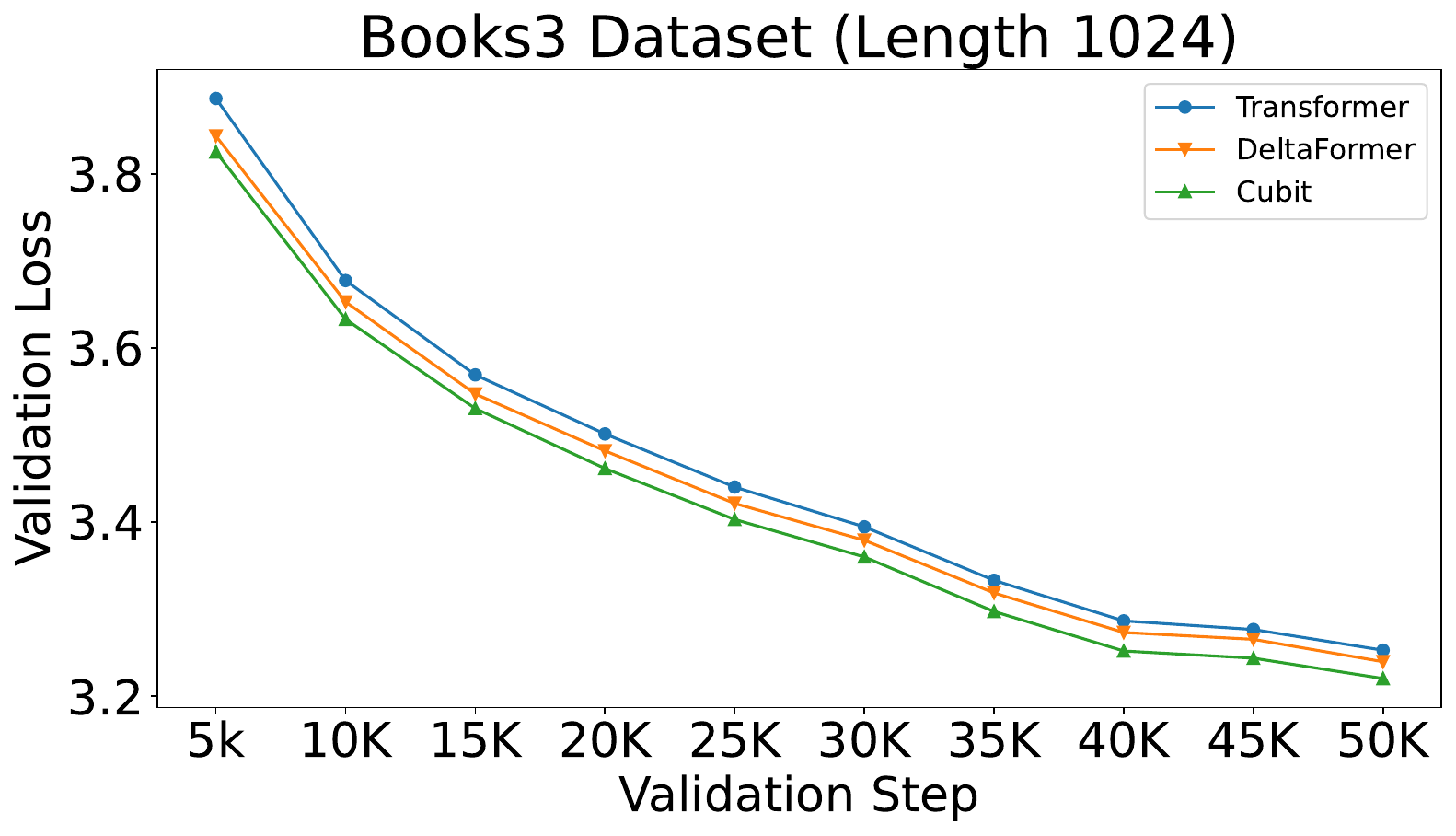}
\caption{
The performance of different methods on the Arxiv and Books3 dataset, with model parameter 125M, training lengths of 512 and 1024.
}
% \vspace{-5pt}
\label{fig: compare with baseline}
\end{figure}

\paragraph{The \methodShort consistently outperforms baselines across diverse datasets.} As shown in Figure \ref{fig: compare with baseline}, our method achieves lower perplexity than both standard Transformer and DeltaFormer on multiple benchmarks. On Arxiv (trained with 512 tokens), \methodShort attains a loss of 1.8802, improving over Transformer (1.9003) and DeltaFormer (1.8983). Similarly, on Books3, it achieves 3.4168, surpassing Transformer (3.4514) and DeltaFormer (3.4371). These results demonstrate \methodShort's robust generalization across different text domains.

\paragraph{The \methodShort scales effectively across sequence lengths.} As shown in Figure \ref{fig: compare with baseline}, our method consistently outperforms the Transformer baseline on both Arxiv and Books3 datasets across 512 and 1024 token training lengths. On Arxiv, \methodShort achieves losses of 1.8802 and 1.7023 versus Transformer's 1.9003 and 1.7210; on Books3, 3.4168 and 3.2204 versus 3.4514 and 3.2529. Notably, the performance gap widens at longer contexts, indicating \methodShort's superior capability in capturing long-range dependencies.

\paragraph{The \methodShort achieves better performance from the beginning of training to the end.} On Books3 dataset With validaton step 5K, the \methodShort achieves 4.0885, which is better than Transformer with 4.1494 loss and DeltaFormer with 4.1044 loss. With 50K validation, the \methodShort sitll achieves better performance than Transformer and DeltaFormer. Smilary, we also observation such result on Arxiv dataset. Therefore, \methodShort achieves better performance from the beginning of training to the end.

\subsection{The Effect of Multi-Domain Datasets}
\begin{figure}[!htbp]
%\begin{figure}
\centering
\includegraphics[width=0.45\textwidth]{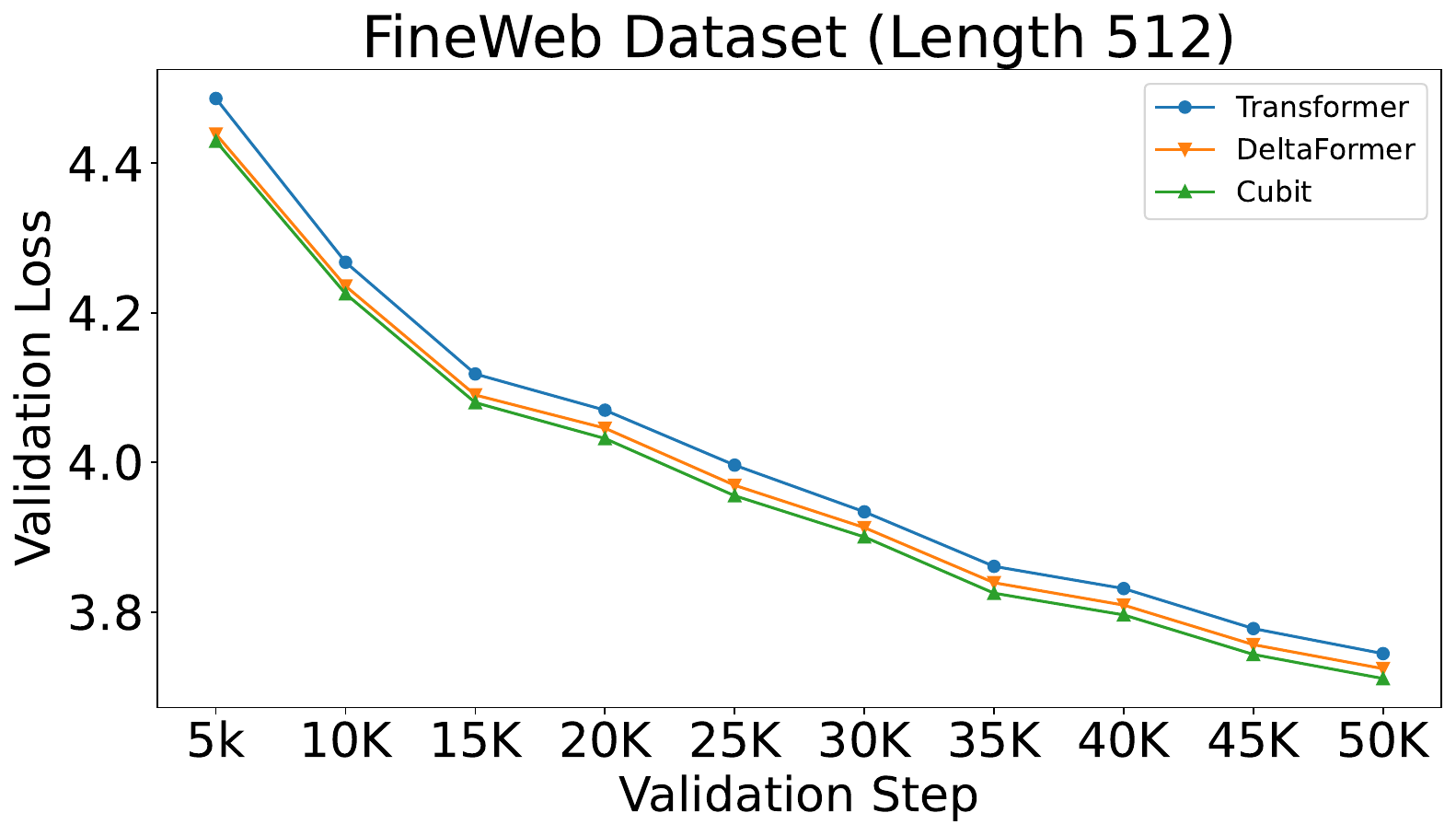}
\hspace{0in}
\includegraphics[width=0.45\textwidth]{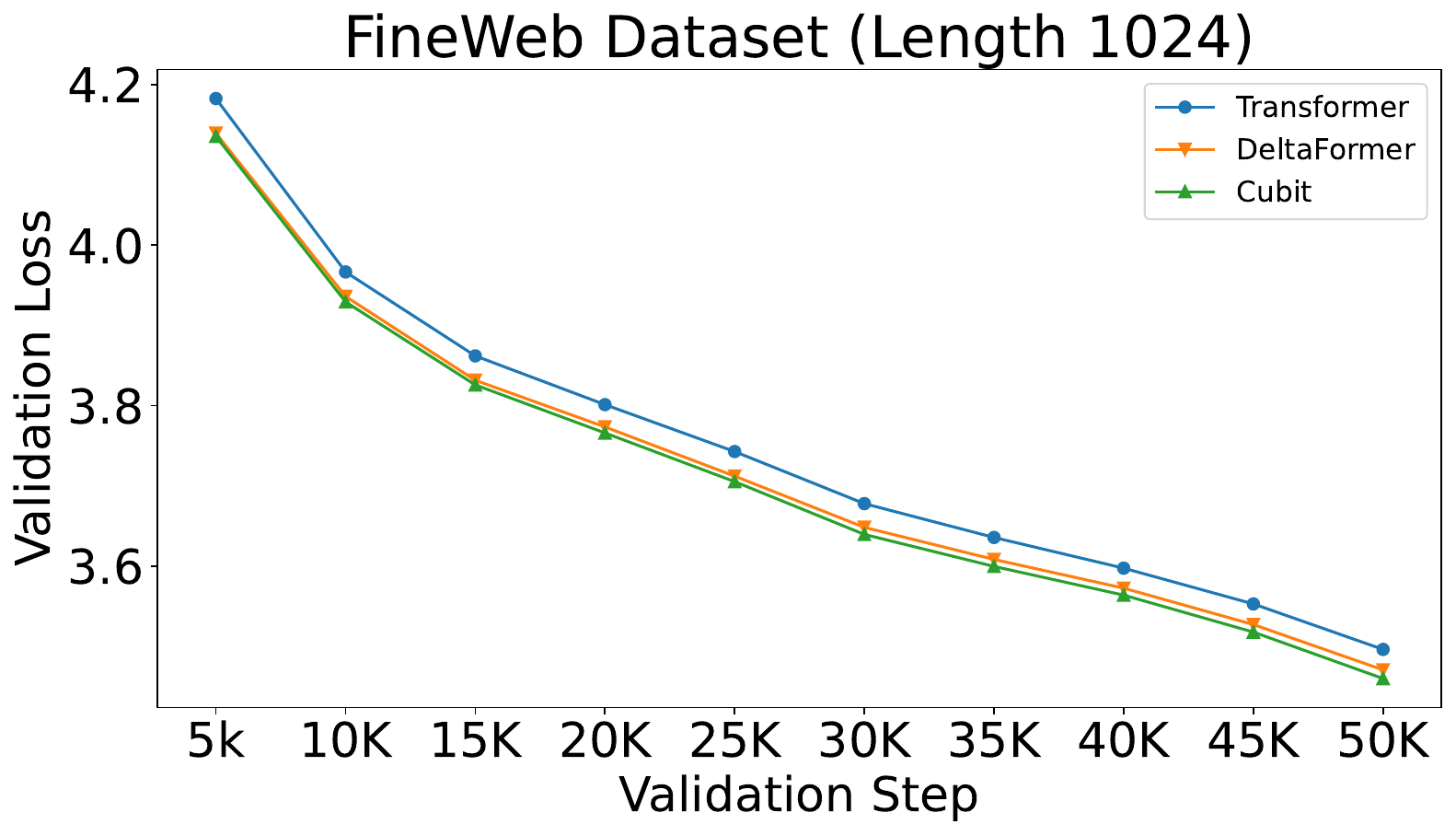}
\caption{
The performance of different methods on the FineWeb dataset, with model parameter 125M.
}
% \vspace{-5pt}
\label{fig: multi_domain}
\end{figure}

\paragraph{The \methodShort achieves better performance on a multi-domain dataset.} As present in Figure \ref{fig: multi_domain}, we evaluate \methodShort against baselines across varying sequence lengths. At 512 tokens, it achieves a validation loss of 3.7110, improving over the Transformer (3.7443) and DeltaFormer (3.7242), respectively. At 1024 tokens, \methodShort maintains its lead with 3.4597 versus 3.4960 (Transformer) and 3.4705 (DeltaFormer). Notably, the performance gap widens with longer sequences, suggesting compounding benefits from its architectural innovations. These consistent improvements across both lengths on a diverse multi-domain dataset establish \methodShort as a potential next-generation architecture for sequence modeling.

\subsection{The Effect of Longer Training Length}
\label{label:training_length}
\begin{figure}[!htbp]
%\begin{figure}
\centering
\includegraphics[width=0.45\textwidth]{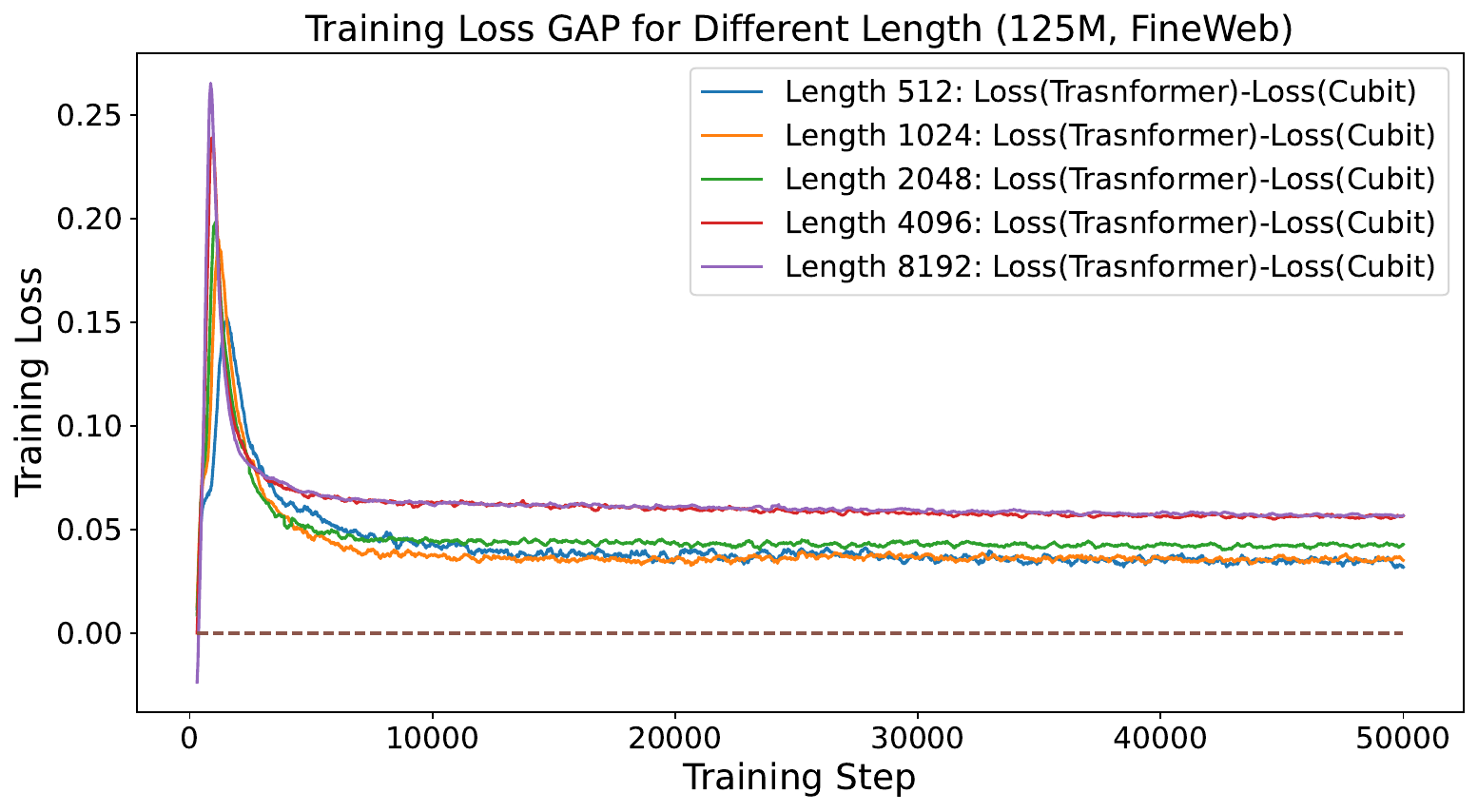}
\hspace{0in}
\includegraphics[width=0.45\textwidth]{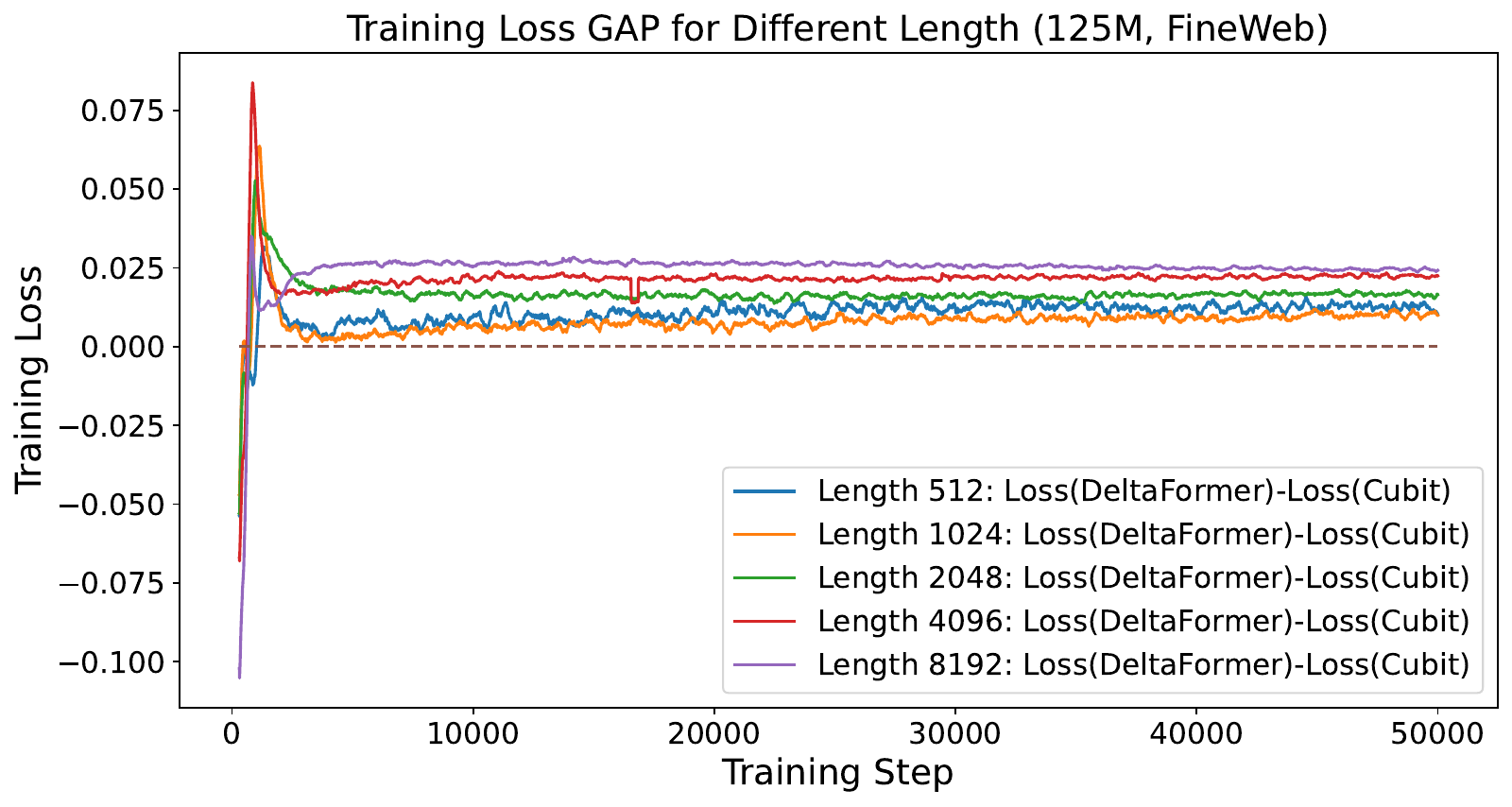}
\caption{
The performance of long training length on the FineWeb dataset, with model parameter 125M.
}
% \vspace{-5pt}
\label{fig: long_length}
\end{figure}

\paragraph{The \methodShort performance gain scales with sequence length.} As presented in Figure \ref{fig: long_length}, training dynamics reveal that \methodShort's advantage over Transformer increases monotonically with sequence length: the loss gap stabilizes at ~0.05--0.06 for 8192 tokens, exceeding margins at 512, 1024 and 4096 tokens. With the training length increase, the loss gap between \methodShort and DeltaFormer also gradually increases. This suggests \methodShort's kernel-based mechanism more effectively captures long-range dependencies, while standard attention suffers at longer contexts.

\paragraph{The gain stems from long-sequence processing, not more tokens.} To disentangle sequence length from total training tokens, we compare Transformer and \methodShort at 1024 tokens with doubled batch size in Figure \ref{fig: training_length_double_batch_size}. The smaller performance gap under increased batch size confirms that improvements derive from architectural efficacy on long sequences rather than simply more training tokens.

\subsection{The Effect of  Large Model}

\begin{figure}[htbp]
%\begin{figure}
\centering
\includegraphics[width=0.45\textwidth]{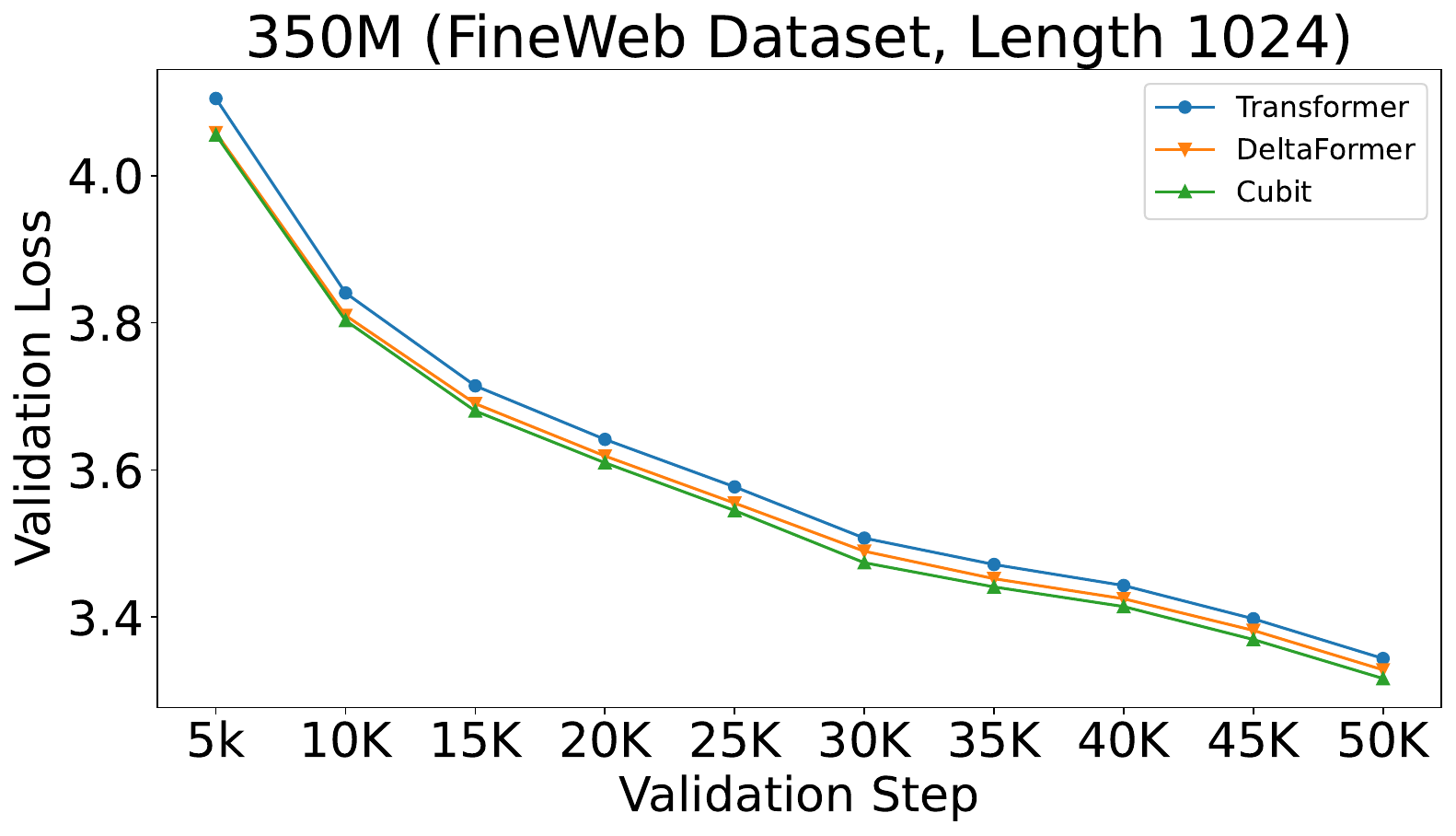}
\hspace{0in}
\includegraphics[width=0.45\textwidth]{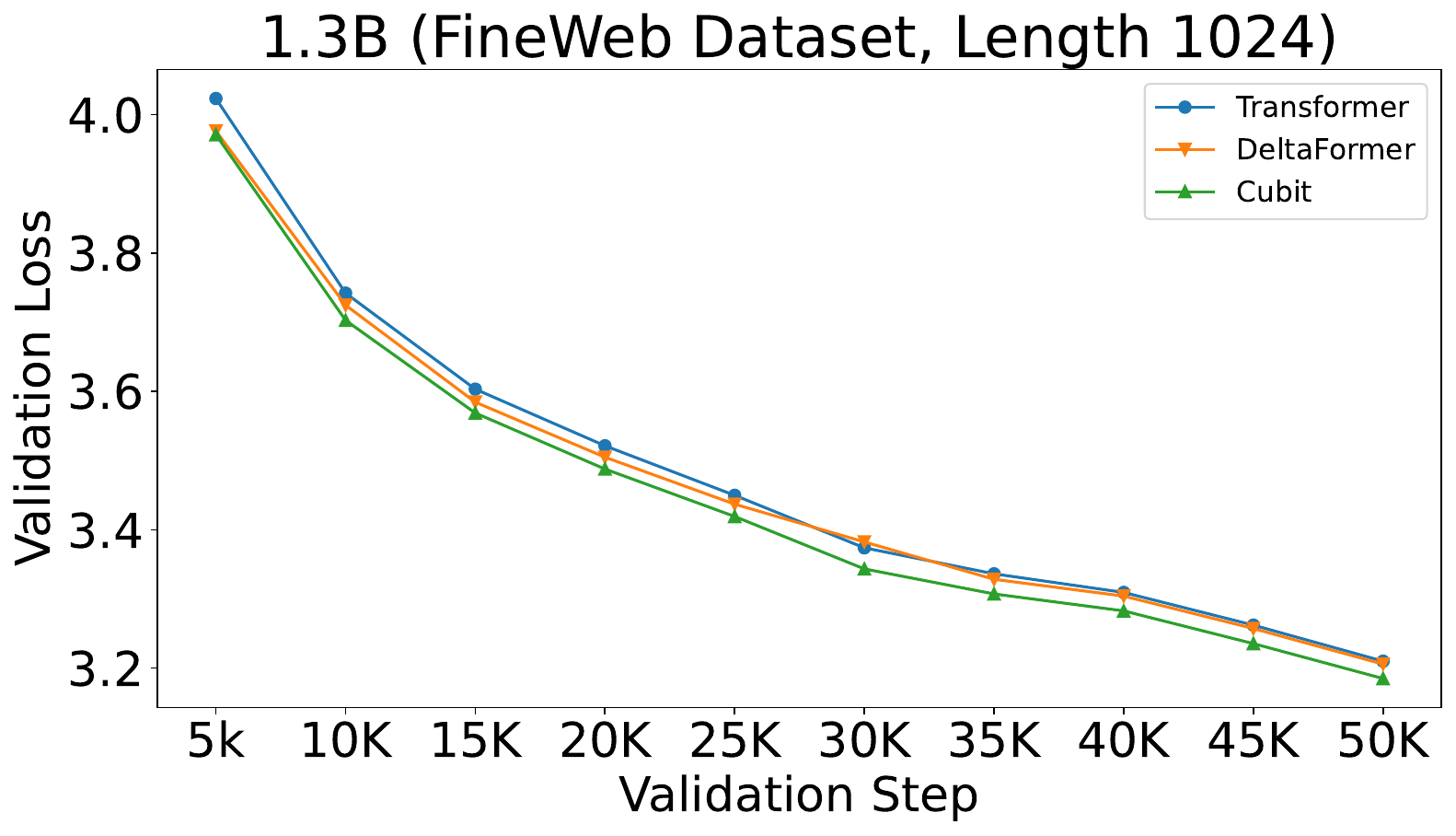}
\caption{
The performance of larger model size on the FineWeb dataset.
}
% \vspace{-5pt}
\label{fig: large_model}
\end{figure}

\paragraph{\methodShort demonstrates favorable scaling properties, whereas DeltaFormer fails to maintain its advantage.} As illustrated in Figure \ref{fig: large_model}, at 350M parameters, \methodShort achieves a loss of 3.3166, substantially outperforming both the standard Transformer (3.3438) and DeltaFormer (3.3284). Upon scaling to 1.3B parameters, \methodShort continues to exhibit strong scaling behavior with a loss of 3.1849, while the performance gap between Transformer (3.2100) and DeltaFormer (3.2057) diminishes considerably. These results suggest that \methodShort benefits consistently from increased model capacity, whereas DeltaFormer's initial improvements over the Transformer baseline erode as model size grows.
\subsection{Ablation Study}
\begin{figure}[!htbp]
%\begin{figure}
\centering
\includegraphics[width=0.45\textwidth]{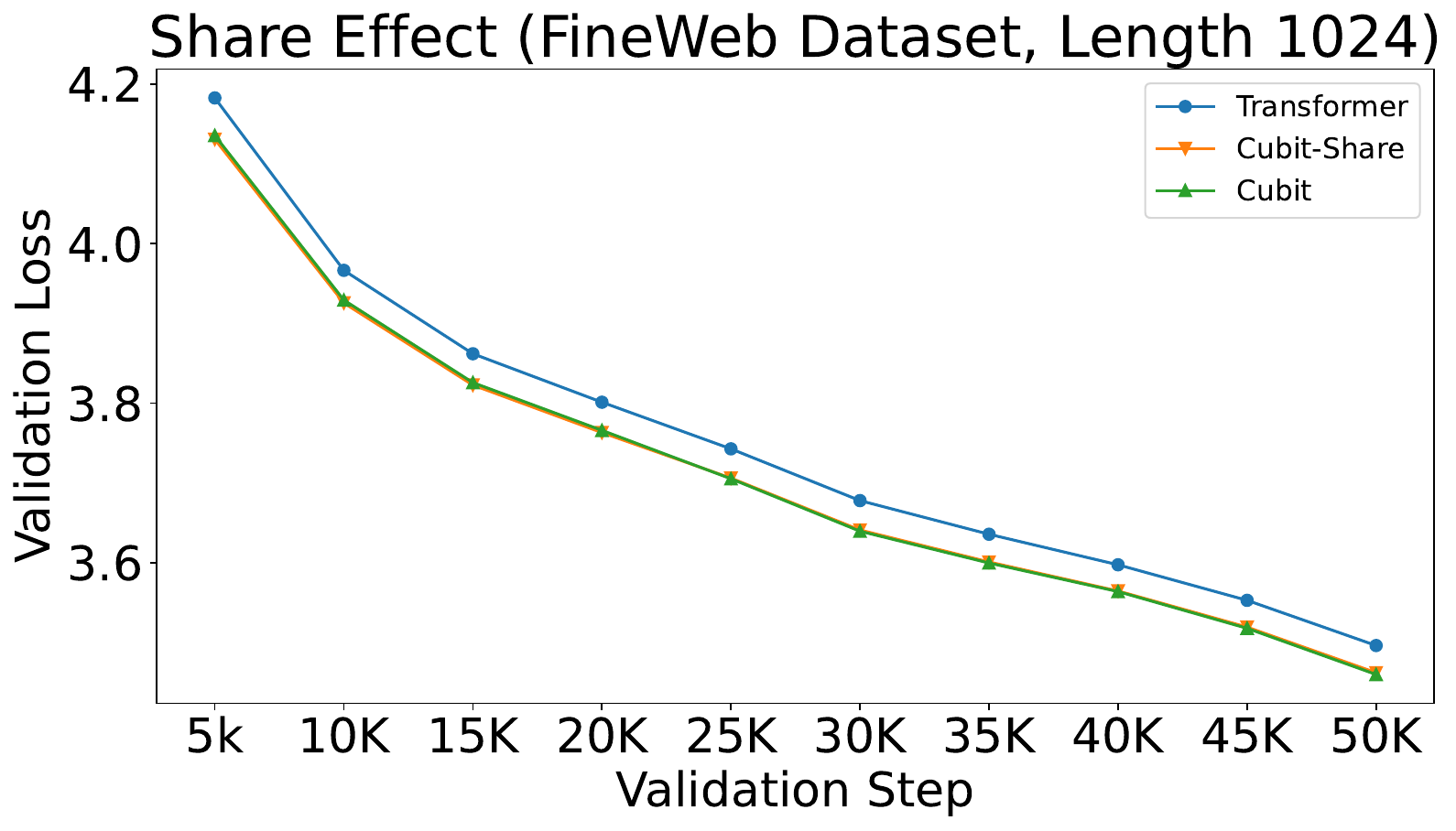}
\hspace{0in}
\includegraphics[width=0.46\textwidth]{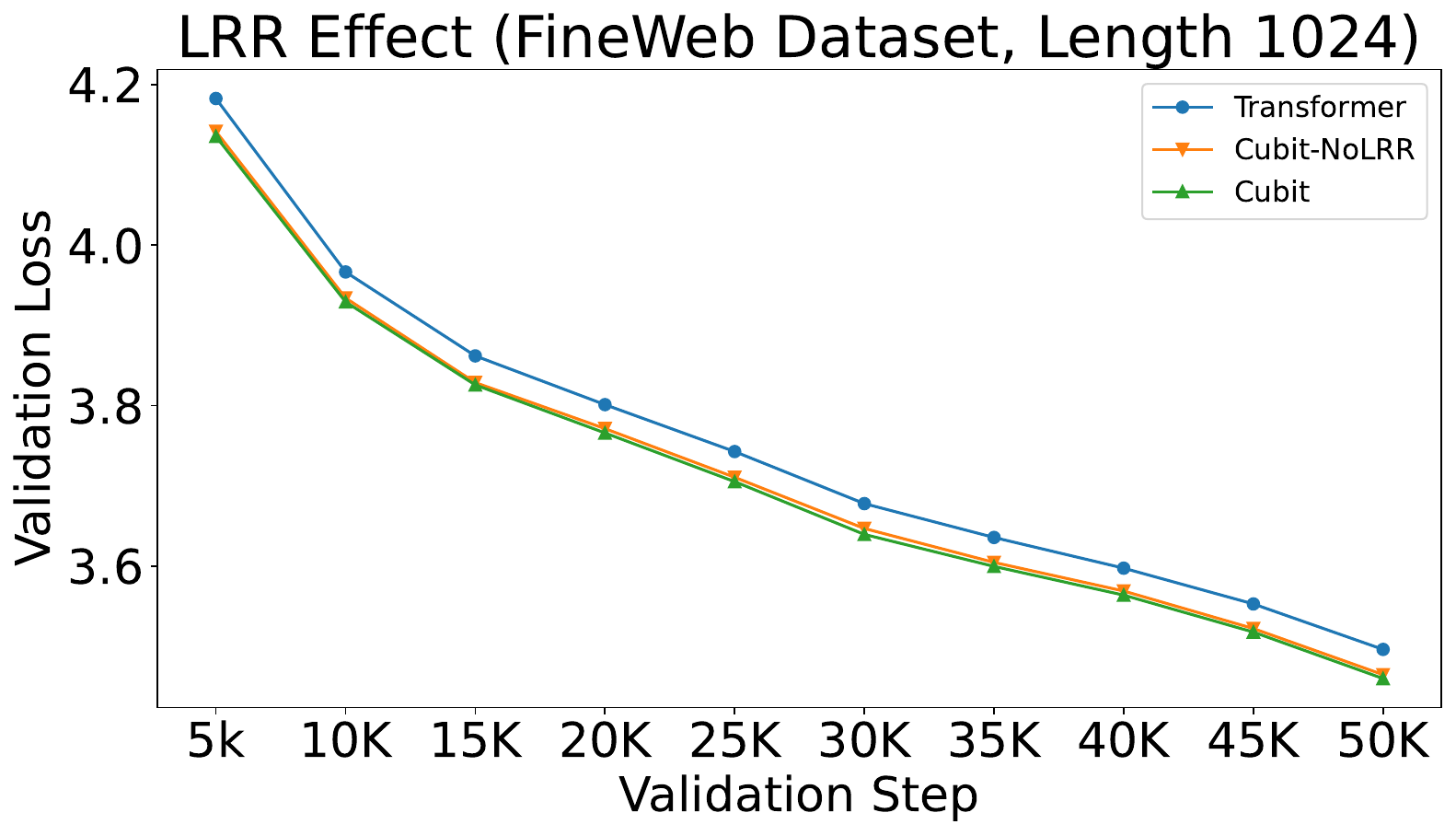}
\caption{
The performance of the share key embedding and no \method, with model size 125M and training length 1024 on the FineWeb dataset.
}
% \vspace{-5pt}
\label{fig: ablation_study}
\end{figure}

\paragraph{Architectural innovation, not parameter expansion, drives performance improvement.} 
Figure \ref{fig: ablation_study} presents an ablation comparing two Reference Embedding configurations: direct Key embedding sharing versus an independent projection matrix. \methodShort equipped with a dedicated projection matrix outperforms both its shared-embedding variant (\methodShort-Share) and the standard Transformer. Critically, even when Reference embeddings are shared, \methodShort maintains consistent superiority over the Transformer across all training stages from initialization through convergence. This establishes that performance gains originate from the structural design rather than increased model capacity.

\paragraph{LRR provides complementary benefits, and the \methodShort-NoLRR is better than Transformer.} 
Incorporating the \methodS mechanism yields a validation loss of 3.4597 for \methodShort, representing a measurable improvement over both the \methodShort-Cubit and the Transformer. Cubit-NoLRR—which operates without the LRR component—achieves a loss of 3.4645, while the conventional Transformer architecture reaches 3.4960 under identical experimental conditions. These results demonstrate that \methodShort is the strongest, followed by Cubit-NoLRR, and Transformer as the baseline.

% \subsection{The \methodShort May Presents Advantages on Longer Length}

% \paragraph{Usually, with the training length increase, the performance gap may gradually become smaller, while such a performance gap is even larger for \methodShort and Transformer.}

\subsection{Performance on Downstream Tasks}

\begin{table}[!htbp]
% \vspace{-10pt}
\caption{Main language modeling results against different methods.
All models are trained on the same subset of the FineWeb-Edu dataset \citep{penedo2024fineweb,lozhkov2024fineweb-edu} with the GPT-2 tokenizer.}
\centering
\resizebox{\textwidth}{!}{
\begin{tabular}{ccccccccc}
\toprule
\textbf{Model}  &     \textbf{ARC-C} &   \textbf{Hellaswag}     & \textbf{PIQA} &  \textbf{SciQ}  &  \textbf{Winograde} &  \textbf{SocialIQA}  & \textbf{RACE} & \textbf{Avg}  \\
\midrule
% {\textit{125M params/10B Tokens }}  \\
% Transformer & 47.35 & 20.48 & 31.59 & 66.21 & 73.60 & 51.85 & 48.51   \\
% DeltaFormer&  49.49 & 20.39 & 34.81 & 69.42 & 75.50 & 49.64 & 49.88   \\
% \methodShort&  51.60 & 21.59 & 35.90 & 69.91 & 76.90 & 53.28 & 51.53  \\
% \midrule
% {\textit{125M params/50B Tokens }}  \\
% Transformer & 47.35 & 20.48 & 31.59 & 66.21 & 73.60 & 51.85 & 48.51   \\
% DeltaFormer&  49.49 & 20.39 & 34.81 & 69.42 & 75.50 & 49.64 & 49.88   \\
% \methodShort&  51.60 & 21.59 & 35.90 & 69.91 & 76.90 & 53.28 & 51.53  \\
% \midrule
{\textit{350M params/50B Tokens}}  \\
Transformer & 45.20 & 44.93 & 69.00 & 69.97 & 53.35 & 32.91 & 31.00 & 49.48   \\
DeltaFormer&  44.82 & 45.52 & 69.80 & 70.57 & 52.57 & 32.91 & 31.20 & 49.63   \\
\methodShort&  45.50 & 45.68 & 68.90 & 71.06 & 54.46 & 32.80 & 31.67 & \textbf{50.01} \\
\midrule
{\textit{350M params/10B Tokens}}  \\
Transformer & 43.10 & 40.82 & 69.60 & 69.42 & 50.28 & 33.06 & 30.62 & 48.13   \\
DeltaFormer&  43.56 & 40.82 & 66.60 & 68.88 & 52.64 & 33.42 & 30.72 & 48.09   \\
\methodShort&  44.02 & 41.54 & 68.20 & 68.44 & 51.14 & 33.73 & 31.10 & \textbf{48.31}  \\
\midrule
{\textit{1.3B params/10B Tokens}}  \\
% \midrule
Transformer & 48.70 & 48.40 & 70.10 & 71.71 & 55.25 & 33.42 & 33.11 & 51.53   \\
DeltaFormer&  47.10 & 48.96 & 70.20 & 72.52 & 54.85 & 32.96 & 32.73 & 51.33   \\
\methodShort&  47.10 & 49.79 & 72.60 & 71.65 & 54.22 & 33.67 & 33.40 & \textbf{51.78}  \\
\midrule
\end{tabular}
}
% \vspace{-10pt}
\label{table: downstream}
\end{table}

\textbf{Downstream Evaluation.}
We evaluate performance on standard benchmarks, including ARC~\citep{clark2018think},  HellaSwag~\citep{zellers2019hellaswag},  PIQA \citep{bisk2020piqa}, SciQ \citep{welbl2017crowdsourcing}, and   WinoGrade~\citep{sakaguchi2021winogrande}, SocialIQA \cite{sap2019social} and RACE \cite{lai2017race}, using the \texttt{lm-evaluation-harness} \citep{eval-harness} codebase.  We train the model with 50K steps with training length 1024 and training tokens 50B. 
The model sizes are 350M, and 1.3B. 
We display the zero-shot evaluation results of models here in Tables~\ref{table: downstream}.

\paragraph{With the same model size, the \methodShort achieves better performance, from small data size (e.g., 10B) to large data size (e.g., 50B).} With the smallest data size 10B, the \methodShort achieves the best average performance 48.31, which is better than the Transformer with 48.13 average performance and DeltaFormer with 48.09 average performance. With the data size increasing to 50B, the \methodShort achieves the best performance 50.01 average performance, which is better than Transformer with 49.48 average performance and DeltaFormer with 49.63 performance. Therefore, the \methodShort consistently achieves better performance, from smaller data size to large data size.

\paragraph{With the same training data size, the \methodShort is always better than the routers, from small model size (e.g., 350M ) to large model size (e.g., 1.3B).} With the model size 350M model size, the \methodShort achieves 48.31 average performance, which is better than Transformer with 48.13 average performance and DeltaFormer with 48.09 average performance. With the model size increase from 350M to 1.3B, the \methodShort achieves 51.78 average performance, which is better than Transformer with 51.53 average performance. And \methodShort is better than DeltaFormer with 51.33 average performance.e Therefore, with the same training data size from small model size to large model size, \methodShort is better than Transformer and DeltaFormer.

\section{Conclusion}
In this work, we propose \methodShort, a novel architecture based on Kernel Ridge Regression that replaces the Nadaraya-Watson estimator underlying Transformers. We conduct extensive evaluations across diverse datasets, sequence lengths, and model scales. \methodShort consistently outperforms the Transformer, with performance gains that become increasingly pronounced at longer training lengths. We believe \methodShort has the potential to be next-generation foundation model architecture.

\nocite{*}

\bibliographystyle{plain}
\bibliography{references}

%%%%%%%%%%%%%%%%%%%%%%%%%%%%%%%%%%%%%%%%%%%%%%%%%%%%%%%%%%%%

\newpage

\appendix

\section{Limitation}
\label{appendix: limitation}
We build \methodShort to replace Transformer, which is validated with sufficient experiments. As \methodShort is more powerful, it is not clear whether \methodShort will be abused.

\section{Broader Impacts}
\label{appendix: broader_impact}
The \methodShort provides a more powerful model, which may benefit society. However, it is not clear whether the \methodShort will be abused.

\section{LLM Usage}
\label{appendix: llm_usage}
In this work, the LLM is used to polish the paper to improve the writing, including the grammatical structures, spelling, punctuation and clarity.

\section{Model Configuration}
\label{model configuration details} 
\begin{table}[htbp]
    \centering
    \setlength{\tabcolsep}{3pt}
    \label{model configuration}
    \caption{\textbf{Model Configurations.}}
    \resizebox{0.6\textwidth}{!}{
    \begin{tabular}{c c c c c}
    \toprule
    & & \textbf{125M} & & \textbf{350M} \\ \midrule
    Training sequence length & & $512$ & & $512$\\
    Batch size & & 32  & & 32 \\
    Number of iterations & & $50$k & & $50$k \\
    Dropout prob. & & $0.0$ & & $0.0$ \\
    Attention dropout prob. & & $0.0$ & & $0.0$ \\
    Attention head && 12 && 16  \\
    Feature dimension && 768 && 1024\\
    Layer number && 12 && 24 \\
    Optimizer & & Adam & & Adam\\
    Optimizer parameter betas & & [0.9, 0.95] && [0.9, 0.95] \\
    Learning rate & & $6\mathrm{e}-4$  & & $3\mathrm{e}-4$ \\
    Precision & & float32 & & float32 \\ 
    \bottomrule
    \end{tabular}
    }
    \label{tab:model_configs}
\end{table}

\section{Local Linear Regression}
\label{appendix: llr}

\begin{table}[!htbp]
% \vspace{-10pt}
\caption{The training loss with different methods, with training length 512 and Books3 dataset}
\centering
\resizebox{\textwidth}{!}{
\begin{tabular}{cccccccccccc}
\toprule
\textbf{Model}  & Regression & 5K  &  10K &   15K  &  20K &25K  &  30K &35K  &  40K &45K  &  50K   \\
\midrule
Transformer & Nadaraya-Watson Regression &4.1948 & 3.9584 & 3.8535 & 3.7920 & 3.7129 & 3.6492 & 3.6007 & 3.5323 & 3.4990 & 3.4701  \\
Cubit & Local Linear Regression & 4.1647 & 3.9389 & 3.8342 & 3.7716 & 3.6959 & 3.6318 & 3.5832 & 3.5155 & 3.4824 & 3.4543 \\
Cubit & Kernel Ridge Regression & 4.1338 & 3.9173 & 3.8141 & 3.7515 & 3.6760 & 3.6118 & 3.5639 & 3.4966 & 3.4643 & 3.4350 \\
\midrule
\end{tabular}
}
\label{table: llr}
\end{table}

The boundary effect constitutes a fundamental challenge in non-parametric regression, where kernel-based estimators exhibit inflated bias near the domain boundaries. Local Linear Regression (LLR) mitigates this phenomenon, revealing complementary perspectives on adaptive smoothing.
Consider the standard NW estimator as a local constant approximation. At a boundary point $\bm{x}$, the asymmetric neighborhood induced by the kernel truncation yields biased estimates because the conditional expectation varies significantly across the effective support. Mathematically, the NW bias scales as $\mathcal{O}(h^2)$ in the interior but deteriorates to $\mathcal{O}(h)$ near boundaries, where $h$ denotes the effective bandwidth \cite{horbunov2024consistency,cheruiyot2020local,jo1997improvement}.

Local Linear Regression addresses this deficiency by fitting a local affine model rather than a constant. The LLR estimator solves:
\begin{equation}
    \min_{\bm{\beta}_0, \bm{\beta}_1} \sum_{j=1}^N K_{ij}^{(h)} \left\| \bm{v}_j^{(h)} - \bm{\beta}_0 - \bm{\beta}_1^\top (\bm{x}_j - \bm{x}_i) \right\|_2^2,
\end{equation}
yielding the prediction $\bm{z}_i^{\text{LLR}} = \bm{\beta}_0^*$. The inclusion of the linear term $\bm{\beta}_1$ enables automatic bias correction: the estimator adapts to local trends, reducing boundary bias to $\mathcal{O}(h^2)$ throughout the domain. In matrix form:
\begin{equation}
    \bm{Z}^{\text{LLR},(h)} = \bm{S}^{(h)} \bm{V}^{(h)}, \quad \text{where} \quad \bm{S}^{(h)} = \bm{e}_1^\top \left(\bm{X}_i^{\top} \bm{W}_i \bm{X}_i\right)^{-1} \bm{X}_i^{\top} \bm{W}_i,
\end{equation}
with $\bm{X}_i = [\bm{1}, (\bm{x}_j - \bm{x}_i)_{j=1}^N]$ the design matrix and $\bm{W}_i = \text{diag}(K_{i1}^{(h)}, \ldots, K_{iN}^{(h)})$ the kernel weights. For the naive implementation without a kernel, the local linear regression is very slow, though the performance may be better because of reducing the boundary effect. For future work, we may use better regression method, such as Local Kernel Ridge Regression \cite{han2022local}. The result is presented in Table \ref{table: llr}.

\paragraph{Future Work.} For this work, we rethink the token mixer with regression, so that we propose Cubit with Kernel Ridge Regression to replace Transformer with Local Linear Regression. And we also analyze the performance of Cubit with Local Linear Regression. In the future, we may propose better regression methods to improve the performance, such as Local Kernel Ridge Regression \cite{han2022local}. And we should also think about whether there is any other theoretical framework that we could use to explain the token mixer. Also, the Local Linear Regression is relatively slow, compared to Cubit with Kernel Ridge Regression. Therefore, in the future, we may also consider how to speed up the Cubit with Local Linear Regression and Kernel Ridge Regression.

\section{The Effect of Training Length and Training Batch Size}
\label{appendix: training_length_batch_size}
\begin{figure}[htbp]
%\begin{figure}
\centering
\includegraphics[width=0.7\textwidth]{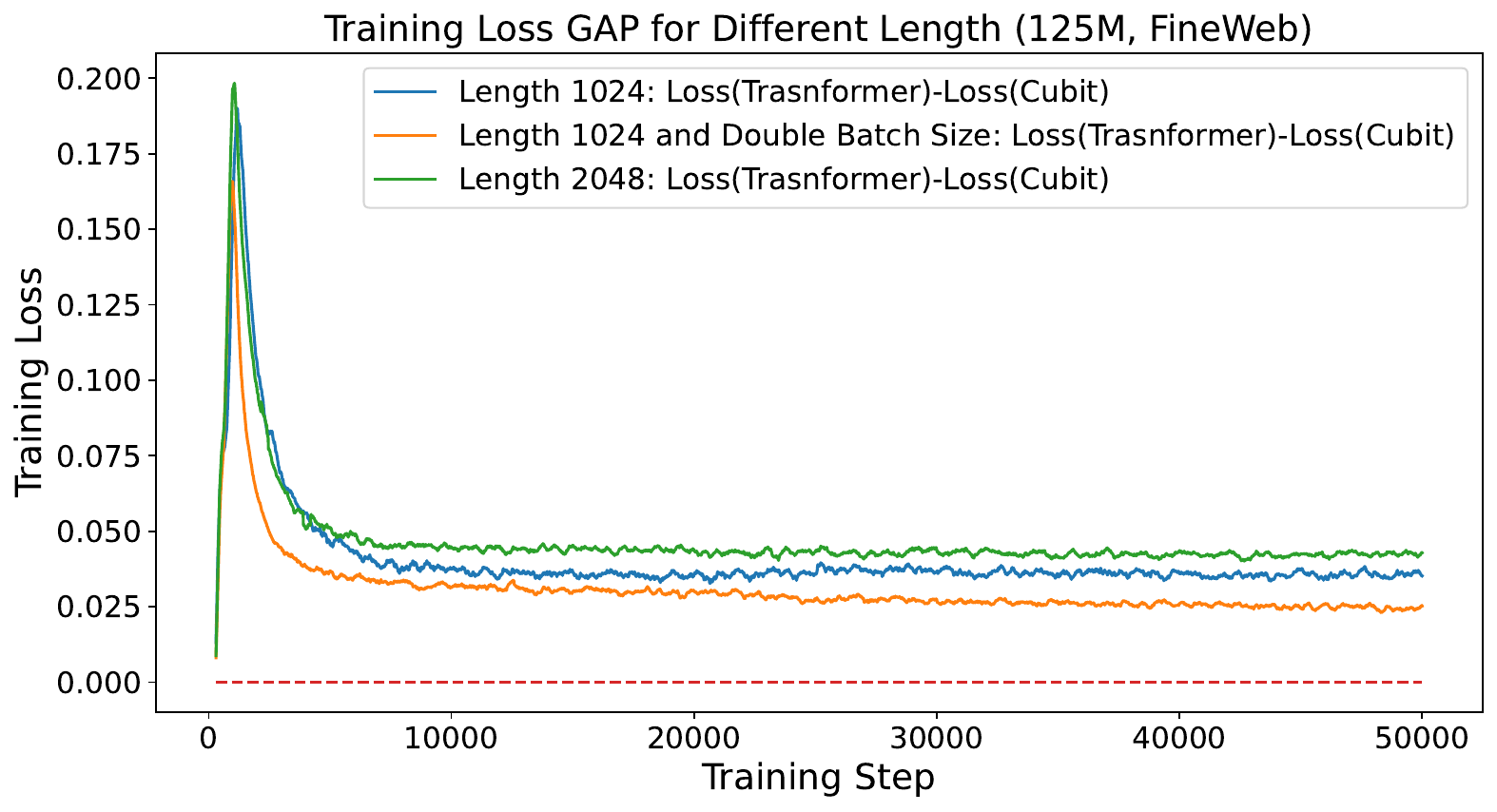}
\caption{
The performance of long training length on the FineWeb dataset, with model parameter 125M. The double batch suggests that the training with double batch size, compared to others.
}
% \vspace{-5pt}
\label{fig: training_length_double_batch_size}
\end{figure}

\newpage

\section{Experiment statistical significance}
\label{appendix: statistical_significance}
\begin{table}[!htbp]
% \vspace{-10pt}
\caption{The validation loss with three random seeds, with training length 1024 and FinWeb dataset}
\centering
\resizebox{\textwidth}{!}{
\begin{tabular}{cccccccccccc}
\toprule
\textbf{Model}  &  & 5K  &  10K &   15K  &  20K &25K  &  30K &35K  &  40K &45K  &  50K   \\
\midrule
Transformer & Mean &4.1768 & 3.9574 & 3.8677 & 3.8142 & 3.7396 & 3.6882 & 3.6409 & 3.5954 & 3.5275 & 3.5185  \\
 & variance & 0.0173 & 0.0077 & 0.0074 & 0.0120 & 0.0070 & 0.0111 & 0.0038 & 0.0116 & 0.0318 & 0.0203 \\
DeltaFormer & Mean &4.1429 & 3.9340 & 3.8471 & 3.7943 & 3.7174 & 3.6674 & 3.6228 & 3.5797 & 3.5114 & 3.5024  \\
 & variance & 0.0106 & 0.0017 & 0.0111 & 0.0198 & 0.0126 & 0.0137 & 0.0112 & 0.0086 & 0.0248 & 0.0239 \\
 \methodShort & Mean &4.1279 & 3.9159 & 3.8293 & 3.7759 & 3.7000 & 3.6493 & 3.6037 & 3.5596 & 3.4913 & 3.4811  \\
 & variance & 0.0122 & 0.0096 & 0.0051 & 0.0123 & 0.0080 & 0.0091 & 0.0030 & 0.0110 & 0.0319 & 0.0195 \\
\midrule
\end{tabular}
}
\label{table: experiment_statistical}
\end{table}

\section{RBF Kernel}

\begin{table}[!htbp]
% \vspace{-10pt}
\caption{Validation loss comparison across different kernel methods and model architectures. We set $\gamma=1/\sqrt{8D}$ to make the variance of $ \gamma | x_i-x_j |^{2} $ be 1, with the assumption that the $x_i$ and $x_j$ elements are normally distributed with variance 1 and mean 0. RBF kernel: $RBF(x_i,x_j)=exp(-\gamma|x_i-x_j|^{2})$. NW: Nadaraya-Watson Regression. KRR: Kernel Ridge Regression}
\centering
\resizebox{\textwidth}{!}{
\begin{tabular}{cccccccccccccc}
\toprule
\textbf{Kernel} & \textbf{Regression} & \textbf{Kernel(QK)} & \textbf{Kernel(KK)} & \textbf{5K} & \textbf{10K} & \textbf{15K} & \textbf{20K} & \textbf{25K} & \textbf{30K} & \textbf{35K} & \textbf{40K} & \textbf{45K} & \textbf{50K} \\
\midrule
RBF & Transformer(NW) & RBF & - & 4.4837 & 4.2662 & 4.1209 & 4.0741 & 3.9985 & 3.9430 & 3.8665 & 3.8371 & 3.7839 & 3.7508 \\
RBF & Cubit(KRR) & RBF & RBF & 5.9308 & 4.9975 & 4.5893 & 4.3778 & 4.2528 & 4.1703 & 4.0820 & 4.0459 & 3.9860 & 3.9499 \\
RBF & Cubit(KRR) & RBF & RBF & 4.4763 & 4.2527 & 4.1002 & 4.0508 & 3.9709 & 3.9132 & 3.8386 & 3.8095 & 3.7558 & 3.7242 \\
Softmax & Transformer(NW) & Softmax & - & 4.4865 & 4.2676 & 4.1182 & 4.0698 & 3.9963 & 3.9339 & 3.8610 & 3.8313 & 3.7778 & 3.7443 \\
Softmax & Cubit(KRR) & Softmax & Softmax & 4.4295 & 4.2254 & 4.0799 & 4.0319 & 3.9554 & 3.9003 & 3.8251 & 3.7961 & 3.7432 & 3.711 \\
\bottomrule
\end{tabular}
}
\label{table:kernel_comparison}
\end{table}

\section{Implementation Details}
\label{appendix: implementation}

In this section, we present the implementation of the proposed \methodShort module in \texttt{PyTorch} \cite{paszke2019pytorch}.

\definecolor{lightgreen}{rgb}{0,0.8,0}
\definecolor{darkgreen}{rgb}{0,0.8,0.2}
\definecolor{backcolour}{rgb}{0.97,0.97,0.94}
\lstset{language=Python,
basicstyle=\smaller\ttfamily,
breaklines=true,
showstringspaces=false,
backgroundcolor = \color{backcolour},
keywordstyle=\color{blue}\ttfamily,
stringstyle=\color{lightgreen}\ttfamily,
commentstyle=\color{gray}\ttfamily,
xleftmargin=2.5em,xrightmargin=0.5em, aboveskip=1em,
morecomment=[l][\color{darkgreen}]{\#}}

\begin{lstlisting}[language=Python]
import torch
import torch.nn as nn
import torch.nn.functional as F
import math

### This is the Cubit with Kernel Ridge Regression, which is the default choice of Cubit.
class Cubit(nn.Module):
    def __init__(
        self,
        hidden_size: int,
        attention_head: int,
        eps: float = 1e-10,
        upper: float = 2.0,
        lower: float = 0.5,
        share: bool = False,
        causal_mask: bool = True
    ):
        super().__init__()

        self.hidden_size = hidden_size
        self.attention_head = attention_head
        self.hidden_size_per_head = hidden_size // attention_head
        self.share = share
        self.causal_mask = causal_mask
        self.eps = eps

        # Linear projections for Q, K, V
        self.q = nn.Linear(hidden_size, hidden_size)
        self.k = nn.Linear(hidden_size, hidden_size)
        self.v = nn.Linear(hidden_size, hidden_size)

        # Optional separate R projection
        if not share:
            self.r = nn.Linear(hidden_size, hidden_size)
        

        # Learnable bounds for LR scale - shape: (1, attention_head, 1, 1)
        self.lower = nn.Parameter(torch.full((1, attention_head, 1, 1), lower))
        self.upper_scale = nn.Parameter(torch.full((1, attention_head, 1, 1), upper - lower))

        # LR scaling factor per head
        self.LRR = nn.Linear(hidden_size, attention_head)

        # Learnable scale for normalized R - shape: (1, attention_head, 1, 1)
        self.scale = nn.Parameter(torch.ones(1, attention_head, 1, 1))

        # Regularization strength lambda (log space for stability) - shape: (1, attention_head, 1, 1)
        self.log_lambda = nn.Parameter(
            torch.full((1, attention_head, 1, 1), math.log(eps)),
            requires_grad=True
        )



    def forward(self, x, pos_encoding_function,softmax, mask) -> torch.Tensor:
        b, t, d = x.shape
        device = x.device

        # Q, K, V, R projections
        q=self.q(x).reshape(b,t,self.attention_head,self.hidden_size_per_head).permute(0,2,1,3)
        k=self.k(x).reshape(b,t,self.attention_head,self.hidden_size_per_head).permute(0,2,1,3) 
        v=self.v(x).reshape(b,t,self.attention_head,self.hidden_size_per_head).permute(0,2,1,3) 
        # Determine R: separate if share=False, else reuse K
        if self.share:
             r=k
        else:
            r=self.r(x).reshape(b,t,self.attention_head,self.hidden_size_per_head).permute(0,2,1,3)
        # LR factor: sigmoid-scaled between lower and upper
        # lrr_logits shape: (B, n_head, T, 1)
        lrr_logits = self.lrr(x).reshape(b, t, self.attention_head, 1).permute(0, 2, 1, 3)
        # lrr shape: (B, n_head, T, 1)
        lrr = self.lower + self.upper_scale * torch.sigmoid(lrr_logits)




        # Normalize R with learnable scale
        # norm_r shape: (B, n_head, T, head_dim)
        norm_r = r / torch.norm(r, dim=-1, p=2, 
        keepdim=True) * self.scale
        
        # add position encoding
        q,k,r,norm_r=pos_encoding_function(q,k,r,norm_r)

        # Sigma inverse: softmax similarity + lambda * I
        # r @ norm_r.transpose(-2, -1) shape: (B, n_head, T, T)
        sigma_inv = softmax(r @ norm_r.transpose(-2, -1), mask)
        
        # I shape: (1, 1, T, T) -> broadcast to (B, n_head, T, T)
        I = torch.eye(t, device=device).unsqueeze(0).unsqueeze(0)
        lambda_reg = torch.exp(self.log_lambda)  # shape: (1, n_head, 1, 1)
        sigma_inv = sigma_inv + lambda_reg * I

        # Solve linear system for each head
        # rhs shape: (B, n_head, T, head_dim)
        rhs = lrr * v
        

        # solution shape: (B, n_head, T, head_dim)

        if self.casual_mask:
            solution = torch.linalg.solve_triangular(sigma_inv, rhs,upper=False)
        else:
            solution = torch.linalg.solve(sigma_inv, rhs)
        


        # Standard attention with Q, K
        A_weights = softmax(
            q @ k.transpose(-2, -1) / math.sqrt(self.hidden_size_per_head), 
            mask
        )  # shape: (B, n_head, T, T)

        # Final output
        output = A_weights @ solution  # shape: (B, n_head, T, head_dim)
        
        # Reshape back to (B, T, hidden_size)
        output = output.permute(0, 2, 1, 3).reshape(b, t, self.hidden_size)
        
        return output

### This is the Cubit with Local Linear Regression
def llr(
    Q: torch.Tensor,
    K: torch.Tensor,
    V: torch.Tensor,
    attention_mask: torch.Tensor,
    weight_func: callable,
    regularization_eps: float = 1.0,
) -> torch.Tensor:
    """
    Local Linear Regression (LLR): we use the non-centered formulation to avoid the M with shape (B,T,T,dim+1)

    Args:
        Q: Query tensor of shape (batch_size, seq_len, dim)
        K: Key tensor of shape (batch_size, seq_len, dim)
        V: Value tensor of shape (batch_size, seq_len, dim)
        attention_mask: Attention mask tensor
        weight_func: Function to compute attention weights from
            similarity scores, such as softmax
        regularization_eps: Ridge regression regularization coefficient. If regularization_eps becomes positive infinity, LLR degrades to Nadaraya-Watson Regression, which is Transformer.

    Returns:
        Output tensor of shape (batch_size, seq_len, dim)
    """
    batch_size, seq_len, dim = Q.shape
    scaling_factor = dim ** 0.5  # Override with standard scaling

    # Step 1: Compute similarity scores
    # Standard scaled dot-product attention scores
    similarity = torch.bmm(Q / scaling_factor, K.transpose(-2, -1))  # (B, T, T)
    similarity = similarity.unsqueeze(1)
    W = weight_func(similarity, attention_mask)  # (B, 1, T, T)
    W = W.squeeze(1)  # (B, T, T)

    # Step 2: Construct augmented feature matrix M = [1, K]
    # Append constant term for intercept estimation in linear regression
    ones = torch.ones(batch_size, seq_len, 1, device=Q.device, dtype=Q.dtype)
    M = torch.cat([ones, K], dim=-1)  # (B, T, 1+dim)

    # Step 3: Compute weighted Gram matrix H = M^T @ diag(W)  @ M
    # This corresponds to X^T X in ordinary least squares
    H = torch.einsum('bij,bjk,bjl->bikl', W, M, M) 

    # Step 4: Compute weighted moment matrix G = M^T @ diag(W) @ V
    # G[b, i, :, :] = M[b, i, :]^T @ diag(W[b, i, :]) @ V[b, :, :]
    # This corresponds to X^T y in ordinary least squares
    G = torch.einsum('bij,bjk,bjl->bikl', W, M, V)  # (B, T, 1+dim, dim)

    # Step 5: Apply ridge regularization (slope terms only, preserve intercept)
    # Regularization matrix: zero out intercept regularization
    reg_matrix = torch.eye(1 + dim, device=H.device, dtype=H.dtype)
    # Do not regularize the intercept term so that the performance degrades to Nadarray-Watson Regression for regularization_eps becomes positive infinity.
    reg_matrix[0, 0] = 0.0  
    H = H + reg_matrix * regularization_eps

    # Step 6: Solve Regression
    # theta = (M^T W M + lambda*I)^{-1} M^T W V
    regression_coeffs = torch.linalg.solve(H, G)  # (B, T, 1+dim, dim)

    # Step 7: Evaluate local linear model at query positions
    intercept = regression_coeffs[:, :, 0, :]          # (B, T, dim) - theta_0
    slope = regression_coeffs[:, :, 1:, :]             # (B, T, dim, dim) - theta_{1:dim}

    # f(q) = theta_0 + theta^T * q
    output = intercept + torch.einsum('btdo,btd->bto', slope, Q)

    return output



\end{lstlisting}

%%%%%%%%%%%%%%%%%%%%%%%%%%%%%%%%%%%%%%%%%%%%%%%%%%%%%%%%%%%%

\end{document}